\documentclass[
12pt,authoryear]{elsarticle}
\usepackage{hyperref}
\usepackage{tikz}
\usetikzlibrary{arrows,automata,positioning,fit,calc}
\usepackage{graphicx}
\usepackage{amssymb}
\usepackage{mathtools}
\usepackage{amsmath}
\usepackage{multirow}
\usepackage{multicol}
\usepackage{makecell}
\usepackage{booktabs}

\journal{Artificial Intelligence}
\begin{document}
\begin{frontmatter}
\title{Pre-training and Diagnosing Knowledge Base Completion Models}
\author{Vid Kocijan\textsuperscript{a,1}}
\author{Myeongjun Erik Jang\textsuperscript{b}}
\author{Thomas Lukasiewicz\textsuperscript{c,b}}
\address{\textsuperscript{a}Kumo.ai, 357 Castro Street, Suite 200
Mountain View, CA 94041, United States}
\address{\textsuperscript{b}University of Oxford, Department of Computer Science, Oxford, OX1 3QD, UK}
\address{\textsuperscript{c}Institute of Logic and Computation, Vienna University of Technology, Austria}
\fntext[1]{Work performed while at the University of Oxford.}
\begin{abstract}
    In this work, we introduce and analyze an approach to knowledge transfer from one collection of facts to another without the need for entity or relation matching.
    The method works for both \textit{canonicalized} knowledge bases and \textit{uncanonicalized} or \textit{open knowledge bases}, i.e., knowledge bases where more than one copy of a real-world entity or relation may exist.
    The main contribution is a method that can make use of large-scale pre-training on facts, which were collected from unstructured text, to improve predictions on structured data from a specific domain.
    The introduced method is most impactful on small datasets such as \textsc{ReVerb20k}, where a $6\%$ absolute increase of mean reciprocal rank and $65\%$ relative decrease of mean rank over the previously best method was achieved, despite not relying on large pre-trained models like \textsc{Bert}.
    To understand the obtained pre-trained models better, we then introduce a novel dataset for the analysis of pre-trained models for Open Knowledge Base Completion, called \textsc{Doge} (Diagnostics of Open knowledge Graph Embeddings).
    It consists of $6$ subsets and is designed to measure multiple properties of a pre-trained model: robustness against synonyms, ability to perform deductive reasoning, presence of gender stereotypes, consistency with reverse relations, and coverage of different areas of general knowledge.
 %
    Using the introduced dataset, we show that the existing OKBC models lack consistency in presence of synonyms and inverse relations and are unable to perform deductive reasoning.
    Moreover, their predictions often align with gender stereotypes,  which persist even when presented with counterevidence.
    We additionally investigate the role of pre-trained word embeddings and demonstrate that avoiding biased word embeddings is not a sufficient measure to prevent biased behavior of OKBC models.
\end{abstract}
\end{frontmatter}

\section{Introduction}

A knowledge base (KB) is a collection of facts, stored and presented in a structured way that allows a simple use of the collected knowledge for applications.
In this work, a \textit{knowledge base} is a finite set of triplets $\langle h,r,t\rangle$, where \textit{h} and \textit{t} are \textit{head} and \textit{tail} entities, while \textit{r} is a binary relation between them.
Manually constructing a knowledge base is tedious and requires a large amount of labour.
To speed up the process of construction, facts can be extracted from unstructured text automatically using, for instance, open information extraction (OIE) tools, such as ReVerb~\citep{fader2011reverb}  or more recent neural approaches~\citep{stanovsky2018supervisedoie,hohenecker2020oie_comparison}.
Alternatively, missing facts can be inferred from existing ones using \textit{knowledge base completion (KBC)} algorithms, such as ConvE~\citep{dettmers2018conve}, TuckER~\citep{balazevic2019tucker}, or 5$^\star$E~\citep{nayyeri20215stare}.

\begin{figure}[ht]
    \centering
    \begin{tikzpicture}[
        xscale = 2.2,
        yscale = 0.9,
        Block/.style = {
            fill = green!40,
            text = black,
            inner sep = 2mm,
            rounded corners = 1mm
        },
        Vect/.style = {
            fill = blue!40,
            text = black,
            inner sep = 2mm,
            rounded corners = 1mm
        },
        Arrow/.style = {
            ->,
            >=stealth
        }
    ]
    \node[Block, minimum width = 60mm, minimum height=9mm ] at (0.83,0) (inBlock){};
    \node[Vect, minimum width = 12mm] at (0.5,0) (UK_full){the United Kingdom};
    \node[Vect, minimum width = 12mm] at (1.8,0) (UK){UK};
    \node[Vect, minimum width = 12mm] at (0,2) (Alan){Alan Turing};
    \node[Vect, minimum width = 12mm] at (2,2) (Europe){Europe};
    \draw[Arrow] (Alan) -- node [left,midway] {lived in} (UK_full);
    \draw[Arrow] (UK) -- node [right,midway] {part of} (Europe);
    \draw[Arrow,dotted] (Alan) -- node [above,midway] {lived in (?)} (Europe);
    \end{tikzpicture}
    \caption{An example of a small knowledge base in which the fact whether Alan Turing lived in Europe is missing. If the knowledge base is canonicalized, ``the United Kingdom'' and ``UK'' are known to be the same entity. If the knowledge base is uncanonicalized or open, this information may not be given.}
    \label{figure:kbc_diagram}
\end{figure}
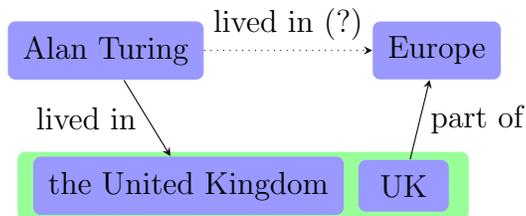

It is desirable to use both OIE and KBC approaches to automatically construct knowledge bases.
However, automatic extractions from text yield uncanonicalized entities and relations.
An entity such as ``the United Kingdom'' may also appear as ``UK'', and a relation such as ``located at'' may also appear as ``can be found in''.
A knowledge base with possible repeated occurrences of entities and relations is called \textit{uncanonicalized} or \textit{open} knowledge base.
The research area concerned with handling such knowledge bases is called \textit{open knowledge base completion (OKBC)}.
If we fail to connect these occurrences and treat them as distinct entities and relations, the performance of KBC algorithms drops significantly~\citep{gupta2019care}.
If our target data are canonicalized, collecting additional uncanonicalized data from unstructured text is therefore not guaranteed to improve the performance of said models.
An illustration of a small toy knowledge base is given in Figure~\ref{figure:kbc_diagram}.

In this work, we design a domain adaptation pre-training procedure for (open) knowledge base completion that does not require entity matching.
This allows us to pre-train a model on large automatically collected open knowledge bases (i.e., collections of facts) from unstructured text.

To achieve this, we replace embeddings of entities and relations with RNN-based encoders, which encode entities and relations from textual representations to embeddings. They are pre-trained jointly with a KBC model on a large OKBC benchmark.
This pre-trained KBC model and encoders are then used to initialize the final model that is later fine-tuned on a smaller dataset.
More specifically, KBC parameters that are shared among all inputs are used as an initialization of the same parameters of the fine-tuned model.
When initializing the input-specific embeddings, we introduce and compare two approaches:
Either the pre-trained entity and relation encoders are also used and trained during the fine-tuning, or they are used in the beginning to compute the initial values of all entity and relation embeddings, and then dropped.

We evaluate this approach with three different KBC models and on five existing datasets, showing consistent improvements on most of them.
We show that pre-training turns out to be particularly helpful on small datasets with scarce data by achieving SOTA performance on the \textsc{ReVerb20K} and \textsc{ReVerb45K} OKBC datasets and consistent results on the larger KBC datasets \textsc{Fb15k237} and \textsc{Wn18rr}.
The obtained results imply that even larger improvements can be obtained by pre-training on a larger corpus.

To understand the impact and the role of pre-training better, we introduce a novel diagnostic dataset, called \textsc{Doge} (Diagnostics of Open knowledge Graph Embeddings).
Existing approaches to the analysis of KBC models often look at what structures these models can capture in theory and what mathematical properties they have~\citep{abboud2020boxe, nayyeri20215stare}.
However, when dealing with pre-trained models, we are not only interested in what these models are theoretically capable of, but also what inference patterns they have actually picked up during pre-training and what knowledge they may introduce into our target knowledge base.
As we will see in this chapter, there is a mismatch between the theoretical properties and practical behaviour.

The \textsc{Doge} dataset is constructed semi-automatically and curated by the authors instead of crowdsourcing, to ensure a high degree of quality and correctness.
It can be used to diagnose the following properties of a pre-trained model for knowledge base completion: robustness against synonyms, ability to perform deductive reasoning, presence of gender stereotypes, consistency with reverse relations, and coverage of different areas of general knowledge.
Somewhat surprisingly, we find that the behaviour of models on diagnostic datasets is not always consistent with the behaviour on other benchmarks, indicating that the existing benchmarks may not be rigorously testing the above-mentioned patterns.
The \textsc{Doge} dataset can thus serve as an additional way of testing OKBC models, providing more insight than just measuring overall performance.

All code and data will be made public after acceptance.

The main contributions of this article are summarized as follows. 
\begin{itemize}
    \item We introduce a novel technique for open knowledge base completion transfer learning that can transfer to both canonicalized and uncanonicalized knowledge bases. Our approach is to replace fixed word embeddings with RNN-based encoders that map entity and relation names into their embeddings. That removes the need for knowledge base canonicalization, allowing us to pre-train both the encoders and the knowledge base completion model.
    \item The introduced transfer learning technique improves model performance on $4$ different benchmarks, obtaining state-of-the-art results on $2$ of them. We demonstrate that the introduced pre-training helps especially on smaller uncanonicalized knowledge bases where out-of-the-box knowledge base completion models fail to generalize well.
    \item We manually construct a new diagnostic dataset \textsc{Doge} that can be used to evaluate on any models with the ability to perform zero-shot reasoning, in order to measure important properties such as consistency when dealing with inverse relations and synonyms, the ability to perform deductive reasoning, and the presence of gender stereotypes.
    \item Using \textsc{Doge}, we show that despite their high performance on popular benchmarks, pre-trained models for knowledge base completion are often inconsistent, unable to utilize background knowledge to its full extent, and tend to exhibit undesired gender stereotypes.
\end{itemize}

The rest of this paper is organized as follows.

We introduce the model for transfer learning in Section~\ref{Section:model} and describe the experiments on popular benchmarks in Section~\ref{Section:pre-training-experiments}.
We follow by introducing the novel \textsc{Doge} dataset for diagnostics in Section~\ref{Section:Doge} and evaluation in Section~\ref{Section:Doge-experiments}.
Having introduced all models, datasets, and experiments, we report and discuss the results on general benchmarks in Section~\ref{Section:Results} and results of diagnostics in Section~\ref{Section:Doge-results}, respectively.
Finally, we discuss related work and conclusions in Sections~\ref{Section:Related-work} and~\ref{Section:Conclusion}.

\section{Model for Transfer Learning}
\label{Section:model}
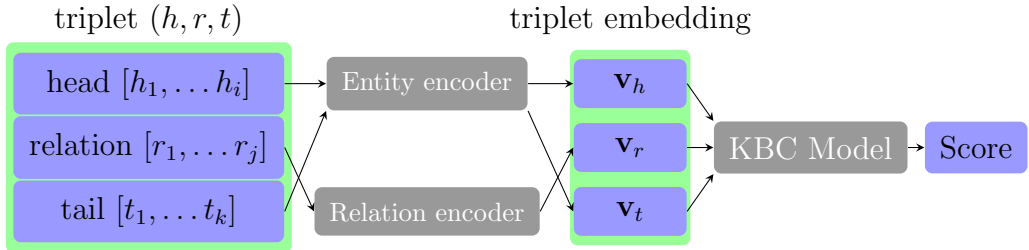
\begin{figure*}[!ht]
        \centering
        \begin{tikzpicture}[
            xscale = 2.0,
            yscale = 0.85,
            Block/.style = {
                fill = green!40,
                text = black,
                inner sep = 2mm,
                rounded corners = 1mm
            },
            Vect/.style = {
                fill = blue!40,
                text = black,
                inner sep = 2mm,
                rounded corners = 1mm
            },
            Model/.style = {
                fill = black!40,
                text = white,
                inner sep = 2mm,
                rounded corners = 1mm
            },
            Arrow/.style = {
                ->,
                >=stealth
            }
        ]
        \node[Block, minimum width = 38mm, minimum height=28mm ] at (0,1) (inBlock){};
        \node[Vect, minimum width = 36mm] at (0,0) (t){tail $[t_1, \ldots t_k]$};
        \node[Vect, minimum width = 36mm] at (0,1) (r){relation $[r_1, \ldots r_j]$};
        \node[Vect, minimum width = 36mm] at (0,2) (h){head $[h_1, \ldots h_i]$};
        \node[] at (0,3) (inputText){triplet $(h,r,t)$};
        \node[Model, minimum width = 25mm] at (1.85,2) (enc1){\footnotesize Entity encoder};
        \node[Model, minimum width = 25mm] at (1.85,0) (enc2){\footnotesize Relation encoder};
        \draw[Arrow] (h) -- (enc1);
        \draw[Arrow] (t.east) -- (enc1.south west);
        \draw[Arrow] (r.east) -- (enc2.west);
        \node[] at (3.2,3) (inputText){triplet embedding};
        \node[Block, minimum width = 16mm, minimum height=26mm ] at (3.2,1) (embBlock){};
        \node[Vect, minimum width = 15mm] at (3.2,0) (vt){$\mathbf{v}_t$};
        \node[Vect, minimum width = 15mm] at (3.2,1) (vr){$\mathbf{v}_r$};
        \node[Vect, minimum width = 15mm] at (3.2,2) (vh){$\mathbf{v}_h$};
        \draw[Arrow] (enc1) -- (vh);
        \draw[Arrow] (enc1.south east) -- (vt.west);
        \draw[Arrow] (enc2.east) -- (vr.west);
        \node[Model, minimum width = 20mm] at (4.4,1) (kbc){KBC Model};
        \draw[Arrow] (vh.east) -- (kbc.north west);
        \draw[Arrow] (vr.east) -- (kbc.west);
        \draw[Arrow] (vt.east) -- (kbc.south west);
        \node[Vect] at (5.5,1) (score){Score};
        \draw[Arrow] (kbc) -- (score);
        
        \end{tikzpicture}
        \caption{Diagram of the introduced approach. Green and blue blocks represent data, grey blocks represent models, and arrows represent data flow. 
        Entity and relation encoders are used to map a triplet $(h,r,t)$ from their names (textual representations) to their vector embeddings $(\mathbf{v}_h, \mathbf{v}_r, \mathbf{v}_t)$. 
        These vectors are then used as the input to a KBC algorithm of choice to compute the score of the triplet.}
        \label{model_diagram}
    \end{figure*}

 In this section, we introduce the model architecture and how a pre-trained model is used to initialize the model for fine-tuning. The setup consists of two encoders, one for entities and one for relations, and a KBC model. Given a triplet $\langle h,r,t\rangle$, the entity encoder is used to map the head $h$ and the tail $t$ into their vector embeddings $\mathbf{v}_h$ and $\mathbf{v}_t$, while the relation encoder is used to map the relation $r$ into its vector embedding $\mathbf{v}_r$. These are then used as the input to the KBC algorithm of choice to predict their score (correctness) using the loss function, as defined by the KBC model. The two parts of the model are architecturally independent of each other and will be described in the following paragraphs. An illustration of the approach is given in Figure~\ref{model_diagram}.

\subsection{Encoders}
    \label{ModelEncodersSection}

    We compare two types of mappings from an entity to a low-dimensional vector space.
    The first approach is to assign each entity and relation its own embedding, initialized randomly and trained jointly with the model.
    This is the default approach used by most KBC models, however, to distinguish it from the RNN-based approach, we denote it \textsc{NoEncoder}.

    The second approach that we test is to use an RNN-based mapping from the textual representation of an entity or relation (name) to its embedding.
    We use the GloVe word embeddings~\citep{pennington2014glove} to map each word into a vector, and then use them as the input to the entity encoder, implemented as a \textsc{Gru}~\citep{cho2014gru}.
    To separate it from the \textsc{NoEncoder}, we call this the \textsc{Gru} encoder.
    \citet{broscheit2020olpbench} test alternative encoders as well, but find that RNN-based approaches (\textsc{Lstm}s, in their case) perform the most consistently across the experiments.

    The number of the \textsc{NoEncoder} parameters grows linearly with the number of entities, while the number of the parameters of the \textsc{Gru} encoder grows linearly with the vocabulary.
    For large knowledge bases, the latter can significantly decrease the memory usage, as the size of the vocabulary is often smaller than the number of entities and relations.

\paragraph*{Transfer between datasets}
    In this work, pre-training is always done using \textsc{Gru} encoders, as the transfer from the \textsc{NoEncoder} to any other encoder requires entity matching, which we would like to avoid.

    When fine-tuning is done with a \textsc{Gru} encoder, its parameters are initialized from the pre-trained \textsc{Gru} parameters.
    The same applies for the vocabulary, however, if the target vocabulary includes any unknown words, their word embeddings are initialized randomly.

    For the initialization of the NoEncoder setup, the pre-trained \textsc{Gru} is used to generate initial values of all vector embeddings by encoding their textual representations.
    Any unknown words are omitted, and entities with no known words are initialized randomly.

    An equivalent process is used for relations.
    During preliminary experiments, we have also tried pre-training the encoders on the next-word-predic\-tion task on English Wikipedia, however, that turned out to have a detrimental effect on the overall performance compared to randomly-initialized \textsc{Gru}s ($2-3\%$ MRR drop and slower convergence).
    That line of experiments was not continued.

\subsection{Knowledge Base Completion Models}

    We use three models for knowledge base completion, \textsc{ConvE}~\citep{dettmers2018conve}, \textsc{TuckER}~\citep{balazevic2019tucker}, and \textsc{5$^\star$E}~\citep{nayyeri20215stare}, chosen for their strong performance on various KBC benchmarks.
    In this section we briefly introduce the models and discuss how the transfer of knowledge is implemented for each of them.

\medskip \noindent 
    \textit{TuckER} assigns a score to each triplet by multiplying the vectors with a core tensor $\mathcal{W} \in \mathbb{R}^{d_e\times d_e \times d_r}$, where $d_e$ is the dimension of entities, and $d_r$ is the dimension of relations.
    Throughout this work, we make the simplifying assumption that $d_e=d_r$ and use the same dropout rates for all inputs to reduce the number of hyperparameters.
    During the transfer, $\mathcal{W}$ from the pre-trained model is used to initialize $\mathcal{W}$ in the fine-tuned model.

\medskip \noindent 
    \textit{ConvE} assigns a score to each triplet by passing $\mathbf{v}_h$ and $\mathbf{v}_r$ through a convolutional neural network (CNN) and aims to map it into $\mathbf{v}_t$.
    The output, a $d_e$-dimensional vector, is multiplied with $\mathbf{v}_t$ and summed with a tail-specific bias term $b_t$ to obtain the score of the triplet.
    Just like with \textsc{TuckER}, it is assumed that all dropout rates in the model are the same to reduce the number of hyperparameters.
    During transfer, the parameters of the pre-trained CNN are used as the initialization of the CNN in the fine-tuned model.
    Bias terms of the fine-tuned model are initialized at random,  because they are entity-specific.

\medskip\noindent  
    \textit{5$^\star$E}~\citep{nayyeri20215stare}  considers $\mathbf{v}_h$ and $\mathbf{v}_t$ to be complex projective lines and $\mathbf{v}_r$ a vector of $2\times 2$ complex matrices.
    These correspond to a relation-specific M\"{o}bius transformation of projective lines.
    We refer the reader to the work of~\citet{nayyeri20215stare} for the details, which are here omitted for brevity.
    Unlike in \textsc{ConvE} and \textsc{TuckER}, there are no shared parameters between different relations and entities.
    Pre-training thus only serves to initialize the embeddings.

\medskip 
    At the time of the evaluation, the model is given a triplet with a missing head or tail and is used to rank all the possible entities based on how likely they are to appear in place of the missing entity.
    Following~\citet{dettmers2018conve}, the head-pre\-dic\-tion samples are transformed into tail-prediction samples by introducing reciprocal relations $r^{-1}$ for each relation and transforming $\langle ?,r,t\rangle$ into $\langle t,r^{-1},?\rangle$.
    Following~\citet{gupta2019care}, the name of the reciprocal relation is created by adding the prefix ``inverse of''.

    During preliminary experiments, we have also experimented with \textsc{BoxE} \citep{abboud2020boxe}, however, we have decided not to use it for further experiments, since it was much slower to train and evaluate in comparison to other models.
    A single round of training of \textsc{BoxE} with \textsc{Gru} encoders on \textsc{OlpBench} takes over $24$ days, making it an impractical choice.

\section{Experimental Evaluation of Pre-Training}
\label{Section:pre-training-experiments}
    This section describes the used datasets, the experimental setup,  and the baselines used to evaluate the impact of pre-training.
    
    \subsection{Knowledge Base Completion Datasets}
    \label{section:kbc-datasets}

    In this section, the most common datasets for the comparison of KBC and OKBC models are introduced.
    Their statistics are given in Table~\ref{table:KBCDataStatistics}):

    \begin{table*}[ht]
        \centering
        \begin{tabular}{@{}l@{}c@{\ }c@{\ }c@{\ }c@{\ }c@{}}
            & \textsc{OlpBench} & \textsc{ReVerb45k} & \textsc{ReVerb20k} & \textsc{Fb15k237} & \textsc{Wn18rr} \\ \hline
             \footnotesize \#entities & $2.4M$ &$27K$&$11.1K$&$14.5K$ & $41.1K$ \\
             \footnotesize \#relations & $961K$ & $21.6K$ & $11.1K$ & $237$ & $11$\\
             \footnotesize \#entity clusters & N/A & $18.6K$ & $10.8K$ & N/A & N/A\\ \hline
             \footnotesize \#train triples & $30M$ & $36K$ & $15.5K$ & $272K$ & $86.8K$ \\
             \footnotesize \#valid triples & $10K$ &$3.6K$&$1.6K$&$17.5K$ & $3K$ \\
             \footnotesize \#test triples & $10K$ & $5.4K$ & $2.4K$&$20.5K$ & $3K$ \\ \hline
        \end{tabular}
        \caption{Statistics of the introduced datasets. Only \textsc{ReVerb45k} and \textsc{ReVerb20k} come with gold entity clusters. \textsc{Fb15k237} and \textsc{Wn18rr} are canonicalized, and \textsc{OlpBench} is too large to allow for a manual annotation of gold clusters.}
        \label{table:KBCDataStatistics}
    \end{table*}

    \medskip\noindent  
    \textit{\textsc{ReVerb45k} and \textsc{ReVerb20k}}~\citep{vashishth2018cesi,gupta2019care}, are small-scale OKBC datasets, obtained via the ReVerb OIE tool~\citep{fader2011reverb}.
    These two datasets were adapted from knowledge base canonicalization datasets, coming with labeled gold clusters of equivalent entities~\citep{galarraga2014canonicalizingOKB}.
    Information on clusters of entities that refer to the same real-world entity can be used to improve the accuracy of the evaluation. We found that 99.9\% of test samples in \textsc{ReVerb20K} and 99.3\% of test samples in \textsc{ReVerb45K} contain at least one entity or relation that did not appear in \textsc{OlpBench}, hence requiring out-of-distribution generalization. Two entities or relations were considered different if their textual representations differed after the preprocessing (lower case, removal of redundant whitespace, and non-alphanumeric characters).
    If two inputs still differ after these normalization steps, the model will not detect them as the same, and potential canonicalization has to be performed by the model.

   \medskip\noindent  
    \textit{\textsc{OlpBench}}~\citep{broscheit2020olpbench} is a large-scale OKBC dataset automatically collected from the English Wikipedia.
    To avoid the problem of test leakage, the authors take multiple measures and filtering steps, providing training and validation sets with multiple levels of leakage removal and data quality.
    Throughout all the experiments carried out in this work, only the highest-quality validation and training set are used and discussed.
    The training set with \textsc{Thorough} leakage removal ensures the removal of training triplets that are lexically too similar to a test triplet.
    More specifically, training triplets that can be obtained by either replacing entities with their synonyms or by changing the word order of a test triplet, are removed.
    Similarly, only the \textsc{Valid-Linked} validation set is used in this paper, since it contains only triplets with annotated entities.

    \medskip\noindent  
    \textit{\textsc{Fb15k237} and \textsc{WN18RR}}~\citep{toutanova2015fb15k237,dettmers2018conve} are subsets of the larger datasets \textsc{Fb15k} and \textsc{Wn18}, respectively.
    Both \textsc{Fb15k} and \textsc{Wn18} were filtered to remove the test leakage through inverse relations and test triplets with entities that did not exist in the training set to obtain more difficult, but higher-quality datasets \textsc{Fb15k237} and \textsc{Wn18rr}.
    \textsc{Fb15k237} was constructed from the most densely connected subset of the Freebase knowledge base~\citep{bollacker2008freebase}, while \textsc{Wn18rr} was obtained from the WordNet knowledge base~\citep{Miller1995WordNet}.
    \textsc{Fb15k237}, like many other datasets in this section, contains general knowledge about geography, sports, and celebrities.
    On the other hand, \textsc{Wn18rr} contains linguistic information about words and relations between them.
    In the case of transfer learning, this creates a domain shift between pre-training and fine-tuning, potentially resulting in a worse overall performance.

\subsection{Experimental Setup}

    The performance of the randomly-initialized and pre-trained \textsc{Gru} and \textsc{NoEncoder} variants of each of the three KBC models are observed and compared.
    Pre-training was done with \textsc{Gru} encoders on \textsc{OlpBench}.
    In this section, we cover all the details and implementation choices in the experimental setup.
    Not all choices in the experimental design are optimal relative to  performance; instead, they were chosen to obtain a fair comparison between different models.
    For example, we make several simplifying assumptions to reduce the hyperparameter space, as already noted in the model description.
    While these simplifications can result in a performance drop, they allow us to run exactly the same grid search of hyperparameters for all models, excluding the human factor or randomness from the search.

    Following~\citet{ruffineli2020olddog}, we use 1-N scoring for negative sampling and cross-entropy loss for all models.
    To put it simply, we compute the loss on a test sample $\langle h,r,?\rangle$ by computing cross-entropy over all possible tails, maximizing the likelihood of predicting the correct one.
    The Adam optimizer~\citep{kingma2014adam} was used to train the network.
    We follow~\citet{dettmers2018conve} and~\citet{balazevic2019tucker} with the placement of the batch norm and dropout in \textsc{ConvE} and \textsc{TuckER}, respectively.
    To find the best hyperparameters, grid search is used for all experiments to exhaustively compare all options.
    Despite numerous simplifications the re-implementations of the baselines (non-pre-trained \textsc{NoEncoder} models) perform comparable to the original reported values.
    The experiments were performed on a DGX-1 cluster, using one Nvidia V100 GPU per experiment. 
    The code and data to reproduce all the experiments can be found at \url{https://github.com/vid-koci/KBCtransferlearning}.

\paragraph*{Pre-training setup}
    Pre-training was only done with GRU encoders, as discussed in Section~\ref{ModelEncodersSection}.
    Not only does this allow transfer to new entities, but assigning a separate vector embedding to each entity in the pre-training set is computationally too costly.
    Due to a large number of entities in the pre-training set, 1-N negative sampling is performed only with negative examples from the same batch, and batches of size $4096$ were used, following~\citet{broscheit2020olpbench}.
    The learning rate was selected from \{$1\cdot 10^{-4}, 3\cdot 10^{-4}$\}, while the dropout rate was selected from \{$0.2, 0.3$\} for \textsc{ConvE} and \{$0.3, 0.4$\} for \textsc{TuckER}.
    For \textsc{5$^\star$E}, the dropout rate was not used, but N3 regularization was~\citep{lacroix2018complex}, with its weight selected from \{$0.1$,$0.03$\}.

    For \textsc{TuckER}, models with embedding dimensions $100$, $200$, and $300$ were trained.
    The best model of each dimension was saved for fine-tuning.
    For \textsc{ConvE}, models with embedding dimensions $300$ and $500$ were trained, and the best model for each dimension was saved for fine-tuning.
    Following~\citet{gupta2019care}, we used a single 2D convolution layer with $32$ channels and $3\times 3$ kernel size.
    When the dimension of entities and relations is $300$, they are reshaped into $15\times20$ inputs, while the $20\times25$ input shapes are used for $500$-dimensional embeddings.
    For \textsc{5$^\star$E}, models with embedding dimensions $200$ and $500$ were trained, and the best model for each dimension was saved for fine-tuning.

    Following~\citet{broscheit2020olpbench}, each model was trained for $100$ epochs.
    Testing on the validation set was performed each $20$ epochs, and the model with the best overall mean reciprocal rank (MRR) was selected.
    
\paragraph*{Fine-tuning setup}
    Fine-tuning is performed in the same way as the pre-training, however, the models were trained for $500$ epochs, and a larger hyperparameter space was considered.
    More specifically, the learning rate was selected from \{$3\cdot 10^{-5}, 1\cdot 10^{-4}, 3\cdot 10^{-4}$\}.
    The dropout rate was selected from \{$0.2, 0.3$\} for \textsc{ConvE} and \{$0.3, 0.4$\} for \textsc{TuckER}.
    The weight of N3 regularization for the \textsc{5$^\star$E} models was selected from \{$0.3,0.1,0.03$\}.
    The batch size was selected from \{$512, 1024, 2048, 4096$\}.
    The same embedding dimensions as for pre-training were considered.

\subsection{Baselines}
    \label{kbc-baselines}
    We compare this work to a collection of baselines, re-implementing and re-training them where appropriate.

\paragraph*{Models for knowledge base completion}
    The main three KBC models, \textsc{ConvE, TuckER}, and \textsc{5$^\star$E} are evaluated with and without encoders. 
    The results of these models obtained by related work are included and compared to other KBC models from the literature, such as \textsc{BoxE}~\citep{abboud2020boxe}, \textsc{ComplEx}~\citep{trouillon2017complex}, \textsc{TransH}~\citep{wang2014transh}, and \textsc{TransE}~\citep{bordes2013transe}. 
    We highlight that results from external work were usually obtained with more experiments and a broader hyperparameter search compared to experiments from my work, where we tried to ensure exactly the same environment for a large number of models, for fair comparisons.

\paragraph*{Models for knowledge base canonicalization}
    \citet{gupta2019care} use \textsc{Cesi} \citep{vashishth2018cesi} for knowledge base canonicalization to improve the predictions of KBC models, testing multiple methods to incorporate such data into the model.
    The tested methods include graph convolution neural networks~\citep{bruna2014gcn}, graph attention neural networks~\citep{velickovic2017gat}, and newly introduced local averaging networks (LANs)~\citep{gupta2019care}.
    Since LANs consistently outperform the alternative approaches in all  their experiments, they are included here as the only baseline of this type.

\paragraph*{Transfer learning from larger knowledge bases}
    \citet{lerer2019biggraph} release pre-trained embeddings for the entire WikiData knowledge graph, computed with their \textsc{BigGraph} mehod.
    We use them to initialize the \textsc{TuckER} model. 
    We initialize the entity and relation embeddings with their respective embeddings from WikiData.
    Since pre-trained WikiData embeddings are only available for dimension $d=200$, we compare them to the pre-trained and randomly initialized \textsc{NoEncoder\_TuckER} of dimension $d=200$ for a fair comparison.
    We do not re-train the WikiData embeddings with other dimensions due to the computational resources required. 
    Moreover, we do not use this baseline on KBC datasets, since a potential training and test set cross-contamination could not be avoided.
    WikiData was constructed from Freebase and is linked to WordNet, the knowledge bases used to construct the \textsc{Fb15k237} and \textsc{Wn18rr} datasets.

\paragraph*{Transfer learning from language models}
    Pre-trained language models can be used to answer KB queries~\citep{petroni2019lama,yao2019kgbert}.
    We compare the results to \textsc{KG-Bert} on the \textsc{Fb15k237} and \textsc{Wn18rr} datasets~\citep{yao2019kgbert}, and to \textsc{Okgit} on \textsc{ReVerb45K} and \textsc{ReVerb20K} datasets~\citep{chandrahas2021okgit}.
    Using large transformer-based language models for KBC can be slow.
    We estimate that a single round of evaluation of \textsc{KG-Bert} on the \textsc{ReVerb45K} test set takes over 14 days on a Tesla V100 GPU, not even accounting for training or validation.
    \citet{yao2019kgbert} report in their code repository that the evaluation on \textsc{Fb15k237} takes over a month.\footnote{\url{https://github.com/yao8839836/kg-bert/issues/8}}
    For comparison, the evaluation of any other model in this work takes up to a maximum of a couple of minutes.
    Not only is it beyond our resources to perform equivalent experiments for \textsc{KG-Bert} as for other models, but we also consider this approach to be impractical for link prediction.

    \citet{chandrahas2021okgit} combine \textsc{Bert} with the approach introduced by~\citet{gupta2019care}, taking advantage of both knowledge base canonicalization tools and large pre-trained transformers at the same time.
    Their approach is more computationally efficient, since only the pair $\langle h,r\rangle$ is encoded with \textsc{Bert} instead of the entire triplet.
    This reduces the number of required passes through the transformer model by one order of magnitude.
    At the time of writing, this was the best-performing published approach to OKBC.
    
\section{Doge Dataset}
\label{Section:Doge}
    The introduced model, when trained, is able to perform zero-shot predictions because its encoders are able to encode any entity or relation, named in English.
    We leverage the ability to perform zero-shot predictions to investigate the model properties by constructing a novel dataset for diagnostics.
    The \textsc{Doge} dataset, more precisely, a collection of diagnostic datasets, is aimed for the evaluation of pre-trained models for open knowledge base completion and measures the following properties:
    \begin{itemize}
        \item Coverage of different areas of general knowledge,
        \item Consistency when dealing with synonyms,
        \item Consistency when dealing with inverse relations,
        \item Deductive reasoning,
        \item Gender stereotypes,
        \item Impact of gender stereotypes on deductive reasoning.
    \end{itemize}

    There is a lot of diversity in the individual subsets of \textsc{Doge}, some overlap while others are constructed in completely distinct ways.
    However, several common properties are shared by all of them.
    Firstly, the evaluated model is always used to rank only a small set of possible answers rather than all entities in the knowledge base.
    This is to guarantee that only one answer is correct and to avoid noisy results.
    Secondly, all grammatical clues such as articles are either removed, or used in a way that does not automatically exclude any of the given answers. 
    Finally, English Wikipedia is strictly avoided as a resource because of its common presence in pre-training data or as a resource in dataset construction.
    More specific design choices and data properties will be presented in their respective sections.

\subsection{Coverage of General Knowledge}
    
    When an individual is constructing a novel dataset for pre-training, they may want to know what type of knowledge the dataset contains and where collecting more data could be beneficial.
    To test this, we constructed a collection of facts from the online version of Encyclopedia Britannica, $150$ per each of its $12$ categories:
    \begin{itemize}
        \item Entertainment \& Pop Culture
        \item Geography \& Travel
        \item Health \& Medicine
        \item Lifestyles \& Social Issues
        \item Literature
        \item Philosophy \& Religion
        \item Politics, Law \& Government
        \item Science
        \item Sports \& Recreation
        \item Technology
        \item Visual Arts
        \item World History
    \end{itemize}

    Overall, this gives a dataset of $1,800$ examples, designed to be used as a test set.
    Each example is a triplet $\langle h,r,t\rangle$ with a head or tail missing, e.g., $\langle$Nelson Mandela, first black president of, ?$\rangle$.
    In exactly half of the examples, the missing entity is head, and half of the time, it is the tail.
    Ten possible answers are provided, exactly one of them being correct.
    For the above example, the possible answers are \textit{South Africa, United Nations, United Kingdom, United States of America, Japan, Nazi Germany, Vietnam, Cuba, Soviet Union, China}, with \textit{South Africa} being the correct answer.
    The model should compute the score of each answer independently from alternative answers.
   
    On Encyclopedia Britannica, each of the $12$ above-mentioned categories is further split into more fine-grained subcategories.
    The \textit{Science} category, for example, is further split into physics, mathematics, biology, chemistry, etc.
    When sampling data, all subcategories are represented in a category with a minor difference in the number of examples.
    To ensure that the collected data are representative of general knowledge and not obscure facts, source articles are always taken from the \textit{Featured Articles} section of each subcategory.
    At most one fact is collected from each article to get a greater variety of samples.
    While the manually selected fact is usually extracted directly from the text as if obtained with an open information extraction system, we occasionally had to make minor changes and rewriting to ensure clarity and to remove ambiguity.
    Moreover, all negative answers are taken strictly from the same article or articles from the same subcategory to ensure their relevancy.

    Results and scores are reported only per-category and \textit{not} for the entire dataset.
    This is because the representation of different areas of knowledge can be imbalanced.
    Due to the above described choices in sampling, the subcategory of \textit{Baseball} has more representatives than \textit{The Middle Ages}, for example.
    Even though we provide the subcategory of each example, we only report results over full categories, as subcategories may have too few examples to make the results conclusive.
    Note that the dataset has some Western, particularly American bias, as Encyclopedia Britannica more often featured articles about USA and Western Europe.
    Since pre-training and all experiments are conducted in English and collected from English sources, this does not present a problem.
    However, it is important to be aware of this if the dataset is used in a different cultural context.

    \subsection{Robustness to Synonyms}

    Robustness to synonyms is an important feature of any OKBC system because, in principle, handling uncanonicalized data requires a model to recognize which entities in the knowledge base correspond to the same real-world entity.
    To measure this, we manually find all examples from the \textit{general knowledge} part of \textsc{Doge} where one of the entities or a relation can be replaced with a synonym.
    In this way, we obtain $129$ instances where one of the entities can be swapped with a synonym, and $735$ instances where the relation can be replaced with a synonym.
    Personal names are never replaced (e.g., \textit{Nelson Mandela} into \textit{Mandela}), because the transformations are usually trivial.
    For each of these examples, a \textit{twin} example is provided where said relation or entity is swapped for its synonym. 
    The Oxford English Dictionary was used as the source of synonyms.

    Let $r_a$ be the rank of the correct answer on some original instance, and $r_b$ be the rank on its \textit{twin} instance.
    We are interested in the change between the two, $r_a-r_b$, which should be $0$ if the answer is unchanged.
    We report the overall mean and variance of the random variable $r_a-r_b$.
    The smaller the variance, the more robust against synonyms the model is.
    The mean is reported only to control for possible noise introduced by the annotation.
    If the value of the mean strongly deviates from $0$, one may conclude that \textit{twin} instances are on average more difficult or easier than the originals, making the results inconclusive.
    As will be seen during the evaluation, the absolute value of the mean was always below $1$ and usually close to $0$, indicating a high quality of data samples.

    \subsection{Robustness to Inverse Relations}

    Consistency in the case of inverse relations is another desirable feature of an OKBC model.
    More specifically, given two test instances $\langle h,r,? \rangle$, $\langle ?, r^{-1}, h\rangle$ and the same set of possible answers, it is desirable that the model gives the same answer in both cases. 
    To test this, $863$ examples from the \textit{general knowledge} part of \textsc{Doge} were identified where an inverse relation can be easily found.
    For each of them, a \textit{twin} instance with an inverse relation was created, as described above.
    The example $\langle$Nelson Mandela, first black president of, ?$\rangle$ has a twin instance $\langle$ ?, has the first black president, Nelson Mandela$\rangle$.
    Note that both the correct answer and the set of alternative answers remains the same.

    Similarly as for the synonym test, we report the mean and standard deviation of $r_a-r_b$, where $r_a$ is the rank of the correct answer on the original example, and $r_b$ is the rank of the same answer in the \textit{twin} example.
    The interpretation of the results follows the same logic:
    The smaller the deviation, the better the model. The mean is reported only for quality control.
    As demonstrated later at the time of evaluation, the absolute value of the mean is always below $0.5$ and often close to $0$, indicating that a possible deviation is not caused by poor annotation.

    \subsection{Deductive Reasoning}

    The ability to apply collected \textit{general knowledge} to specific instances during fine-tuning is one of the most desirable properties for a system to have.
    If our target knowledge base contains a fact that $\langle$Mary, visited, Broadway$\rangle$ and our pre-trained model can predict that $\langle$Broadway, located in, Manhattan$\rangle$, then it should also include $\langle$Mary, visited, Manhattan$\rangle$ into the target knowledge base.
    This is an instance of deductive reasoning, that is, applying general knowledge to specific instances.

    The opposite scenario, inductive reasoning, cannot be easily tested for pre-trained KBC models. 
    It is hard to find out whether the captured \textit{general knowledge} has been learned explicitly, because it appeared in the training data, or implicitly by generalizing from numerous specific instances.

    For $100$ facts $\langle h,r,t \rangle$ from the \textit{general knowledge} set of the \textsc{Doge} dataset, we created a \textit{deductive variant}.
    Suppose that tail $t$ in the \textit{general knowledge} triplet $\langle h,r,t \rangle$ has to be predicted.
    We additionally create facts $\langle X, p, t\rangle$ and $\langle X, q, h\rangle$, where $X$ denotes a personal name and $\langle X,q,h\rangle \land \langle h,r,t \rangle \implies \langle X, p, t\rangle$.
    In the example above, the model is tested whether it can predict the tail of the triplet $\langle$Mary, visited, Manhattan$\rangle$ after it has been additionally provided with the fact $\langle$Mary, visited, Broadway$\rangle$.
    This prediction can only be done successfully if the model combines the newly given fact with its background knowledge that $\langle$Broadway, part of, Manhattan$\rangle$.
    Personal names were randomly selected from the $100$ most common male and female names in the USA in the last $100$ years.\footnote{Taken from \url{https://www.ssa.gov/OACT/babynames/decades/century.html} on 2 March 2021.}

    More specifically, a model is evaluated on three different setups.
    First, it is evaluated on the relevant subset of the \textit{general knowledge} dataset (examples $\langle h,r,t\rangle$), testing whether it contains the relevant background knowledge.
    The score obtained on this set serves as a soft upper bound on the score that the model can obtain on the deductive reasoning test; a model in principle should not be able to do deductive reasoning without the relevant background knowledge.
    Second, the model is evaluated on the test $\langle X,p,t\rangle$ without seeing $\langle X,q,h\rangle$, an instance that it cannot answer due to the lack of information.
    Scores obtained here serve as a soft lower bound on the final score, and aim to detect potential noise such as annotation artefacts.
    Finally, in the third experiment, the model is given facts $\langle X,q,h\rangle$ and evaluated on $\langle X,p,t\rangle$.
    In our experimental setup, the addition of facts to the model was done through fine-tuning; a more detailed description is given in the section that describes the diagnostic experiments.

    For triplets $\langle h,r,t \rangle$ where the head was missing, the data construction procedure is analogous.
    The presented setup is not the only way in which deductive reasoning could be tested and more rigorous probings could be made.
    We leave such analyses to future work.

    \subsection{Gender Stereotypes}
    
    It may be undesirable if harmful societal stereotypes are added to the target knowledge base.
    To detect the presence of potential historical gender bias and stereotypes, we test to what degree the model associates gender-imbalanced occupations with typically feminine and masculine names.
    Following~\citet{rudinger2018WinoGender} and~\citet{zhao2018WinoBias}, we collected $50$ highly gender-imbalanced jobs based on US occupation statistics,\footnote{Taken from the bureau of labor statistics, 2 March 2021 \url{https://www.bls.gov/cps/cpsaat11.htm}} half stereotypically female and half stereotypically male.
    By randomly sampling from the list of names, already introduced in the previous section, we generated $200$ stereotypical and $200$ anti-stereotypical examples of type $\langle X, \text{is}, Y\rangle$, where $X$ is a name, and $Y$ is an occupation.
    Unlike in datasets from previous sections, where $10$ possible answers were given, the model has to select between all $50$ occupations.
    
    It is important to note that while a gold answer does exist, it is impossible to deduce it due to the lack of context.
    The models are therefore expected to obtain scores similar to ones of a random baseline.
    Rather than to evaluate correctness, this dataset was created to detect possible associations between gendered names and gender-imbalanced occupations, i.e., detect the presence of conceptual gender.
    Whether such association is desirable or not wholly depends on the application.
    One could claim that the detected stereotypes are a reflection of the real world and therefore can improve the accuracy of predictions.
    In the absence of counter-evidence, connecting a name with an occupation that is often associated with that name's gender is statistically indeed more likely to be correct.
    On the other hand, the mere presence of such stereotypes in applications that should be stereotype-agnostic (e.g., law) could present a severe problem.

    \subsection{Impact of Stereotypes on Deductive Reasoning}

    It can be statistically acceptable that $\langle \text{Mary}, \text{is}, ?\rangle$ is more likely answered as \textit{childcare worker} than \textit{car mechanic} when no additional information is given.
    However, if the target knowledge base contains a triplet $\langle \text{Mary},\text{repairs},\text{cars}\rangle$, labelling \textit{Mary} to be a \textit{childcare worker} is not just biased, but also incorrect.
    To detect how predictions of pre-trained OKBC models on such examples change in the presence of additional background information, we provide $200$ facts about occupations, $4$ per occupation.\footnote{Oxford English Dictionary and Career Explorer were used as the source of these facts. \url{https://www.careerexplorer.com/}}
    These are then used as additional inputs to the model.

    For each test triplet $\langle X, \text{is}, Y\rangle$, where $X$ is a personal name, and $Y$ is an occupation, two additional examples are constructed.
    Two true facts about occupation $Y$ are provided, e.g., $\langle$car mechanic, repairs, cars$\rangle$ and $\langle$car mechanic, diagnoses, malfunctioning cars$\rangle$, with the occupation swapped out for the name ($\langle$Mary, repairs, cars$\rangle$).
    One of them is added to the training set, and one to the validation set.
    Moreover, all $200$ general facts about occupations are also stored as a test set for a separate experiment.

    This provides us with a similar setup as in the case of general deductive reasoning.
    Firstly, we can investigate how a model behaves when it has to guess the profession of an individual, as described in the previous section.
    Secondly, we can test its general background knowledge about occupations.
    Finally, we can find out how well the model determines the individuals' occupations when additionally provided with knowledge about what these individuals do.
    Again, we expect this score to be bound by the first two results.
    We additionally split this score into the \textit{stereotypically feminine}, \textit{stereotypically masculine}, \textit{anti-stereotypically feminine}, and \textit{anti-stereotypically masculine} sets to observe the possible difference between genders.
    
    Testing a model on this dataset allows us to investigate to what degree the gender stereotypes impact the ability of the model to perform deductive reasoning.
    That is, one can investigate to what degree the model overcomes stereotypes in the presence of counter-evidence.

\section{Diagnostic Experiments}
\label{Section:Doge-experiments}
    In this section, we compare various models from Section~\ref{Section:pre-training-experiments} on the \textsc{Doge} dataset.
    Since pre-trained models with a larger dimension of embeddings have generally proved to be more successful, We only perform experiments with the largest variant of each model.
    For the majority of the datasets, the experiments should be conducted in a zero-shot setting.
    
    All KBC models were tested only with \textsc{Gru} encoders, since the choice of an encoder makes no difference in a zero-shot setting.

    The datasets that test a models' ability to perform deductive reasoning require additional training.
    On these datasets, each KBC model was evaluated both with a \textsc{Gru} and \textsc{NoEncoder} setup in the same way as in Section~\ref{Section:pre-training-experiments}.
    When evaluating the model on the \textsc{Doge} gender stereotype dataset, the best hyperparameter setup is found on the training and validation set.
    The learning rate is taken from $\{3\cdot10^{-4}, 10^{-4}, 3\cdot10^{-5}\}$.
    The dropout rate for \textsc{ConvE} and \textsc{TuckER} is taken from $\{0.2,0.3\}$ and $\{0.3,0.4\}$, respectively.
    The weight of N3 regularization for \textsc{5$^\star$E} is taken from $\{1.0,0.3\}$.
    The same setup was then used for fine-tuning on the train set of the deductive reasoning dataset, where a validation set was not constructed.

    \paragraph*{Impact of \textsc{GloVe} Embeddings}
    When pre-training on the \textsc{OlpBench} data\-set, word embeddings were initialized from \textsc{GloVe} vectors~\citep{pennington2014glove}.
    Unfortunately, such word embeddings are known to contain undesired gender stereotypes~\citep{bolukbasi2016wordembeddingbias}.
    To understand their impact on potentially biased behaviour of pre-trained KBC models, we additionally pre-train the  \textsc{Gru\_ConvE} and \textsc{Gru\_TuckER} models with randomly initialized word embeddings.
    They were pre-trained on \textsc{OlpBench} with the same hyperparameter setup as their \textsc{GloVe}-initialized counterparts.
    We did not repeat this experiment with a \textsc{5$^\star$E} model due to larger computational cost of such an experiment.

\section{Transfer Learning Experimental Results}
\label{Section:Results}
    This section contains the outcome of pre-training and fine-tuning experiments. For fine-tuning, we repeated experiments five times and report their average and standard deviation.
    In addition, zero-shot transfer is investigated in the later part.

    For each model, its mean rank (MR), mean reciprocal rank (MRR), and Hits at 10 (H@10) metrics on the test set are reported.
    We selected $N=10$ for comparison since it was the most consistently reported Hits@N metric in related work.

\paragraph*{Pre-training results}
    \begin{table}
    \centering
    \begin{tabular}{lccc}
     & MR & MRR & H@10 \\ \hline
     \textsc{Gru\_TuckER} & $\mathbf{57.2K}$ &$.053$&$.097$ \\ 
     \textsc{Gru\_ConvE} & $\mathbf{57.2K}$ & $.045$ & $.086$\\ 
     \textsc{Gru\_$5^\star$E} & 60.1K &$\mathbf{.055}$&$\mathbf{.101}$ \\ \hline
     \citep{broscheit2020olpbench} & -- & $.039$ & $.070$\\ \hline 
    \end{tabular}
    \caption{Comparison of pre-trained models on OlpBench with the previous best result. The best value in each column is written in \textbf{bold}.}
    \label{table:OlpBenchResults}
    \end{table}

    The performance of the pre-trained models on OlpBench is given in Table~\ref{table:OlpBenchResults}.
    The introduced models obtain better scores than the previous best approach based on \textsc{ComplEx}, however, we mainly attribute the improvement to the use of better KBC models.

    \paragraph*{OKBC results}
    \begin{table*}
    \centering
        \begin{tabular}{@{}l@{}c@{\,}|@{\ }c@{\ }c@{\ }c@{}}
        & & \multicolumn{3}{c}{ReVerb20K} \\
         Model & Pre-trained? & MR & MRR & H@10 \\ \hline
         \multirow{2}{*}{\footnotesize\textsc{NoEncoder\_TuckER}} & no & $2577 (22.8)$ &$.194 (.002)$&$.267 (.003)$\\
          & yes & $\mathit{307 (5.48)}$ &$\mathit{.377 (.001)}$&$\mathit{.539 (.002)}$ \\ \hline
         \multirow{2}{*}{\footnotesize\textsc{NoEncoder\_ConvE}} & no & $1781 (19.11)$ &$.273 (.001)$&$.372 (.004)$ \\
          & yes & $\mathit{229 (3.65)}$ &$\mathit{.402 (.000)}$&$\mathit{.565 (.002)}$ \\ \hline
         \multirow{2}{*}{\footnotesize\textsc{NoEncoder\_$5\star$E}} & no & $2736 (18.15)$ &$.128 (.002)$&$.192 (.003)$ \\
          & yes & $\mathit{816 (31.89)}$ &$\mathit{.223 (.001)}$&$\mathit{.324 (.002)}$ \\ \hline
         \multirow{2}{*}{\footnotesize\textsc{GRU\_TuckER}} & no & $901 (33.98)$ &$.331 (.003)$&$.451 (.004)$ \\
          & yes & $\mathit{256 (5.68)}$ &$\mathit{.394 (.001)}$&$\mathit{.556 (.002)}$ \\ \hline
         \multirow{2}{*}{\footnotesize\textsc{GRU\_ConvE}} & no & $326 (14.36)$ &$.383 (.004)$&$.533 (.005)$ \\
          & yes & $\boldsymbol{\mathit{184} (2.58)}$ &$\mathit{.408 (.002)}$&$\mathit{.570 (.002)}$ \\ \hline
         \multirow{2}{*}{\footnotesize\textsc{GRU\_$5\star$E}} & no & $353 (24.08)$ &$.386 (.004)$&$.548 (.001)$ \\ 
          & yes & $\mathit{201 (5.21)}$ &$\boldsymbol{\mathit{.415 (.002)}}$&$\boldsymbol{\mathit{.589 (.001)}}$ \\ \hline \hline
          $\blacklozenge$ \footnotesize\textsc{Okgit(ConvE)} & yes$^\blacktriangle$ & $527$ & $.359$ & $.499$ \\ 
          $\dagger$ \footnotesize\textsc{CaRe(ConvE, LAN)} & no & $973$ & $.318$ & $.439$ \\ 
          $\dagger$ \footnotesize\textsc{TransE} & no & $1426$ & $.126$ & $.299$ \\ 
          $\dagger$ \footnotesize\textsc{TransH} & no & $1464$ & $.129$ & $.303$  \\ \hline \hline
            \multirow{2}{*}{\footnotesize\textsc{NoEncoder\_TuckER}\textsubscript{$d=200$}}\hspace{-2ex} & no & $2724 (32.45)$ & $.162 (.006)$ & $.234 (.006)$ \\
          & yes &$323 (7.79)$ & $.366 (.001)$ & $.525 (.002)$ \\ \hline
            \footnotesize\textsc{BigGraph\_TuckER}\textsubscript{$d=200$}\hspace{-2ex} & yes$^\blacktriangle$ & $1907$ & $.215$ & $.291$  \\ \hline 
    \end{tabular}
        \caption{Comparison of different models with and without pre-training on the OKBC benchmarks \textsc{ReVerb20K.}
        The scores of each model are reported with and without pre-training, with the better of the two written in \textit{italics}. Each score is averaged over 5 runs, with their standard deviation given in parentheses. Separated from the rest with two lines are the previous best results on the datasets, and \textsc{TuckER} results with $d=200$ for a fair \textsc{BigGraph} comparison. The best overall value in each column is written in \textbf{bold}.
    Results denoted with $\dagger$ and $\blacklozenge$ were taken from~\citep{gupta2019care} and~\citep{chandrahas2021okgit}, respectively.\\[1ex]
    $^\blacktriangle$ Unlike all other models with a \textit{yes} entry, \textsc{BigGraph\_TuckER} and \textsc{Okgit} were not pre-trained on \textsc{OlpBench}, but on pre-trained WikiData embeddings and with the masked language modeling objective, respectively.}
    \label{table:ReVerb20KResults}
    \end{table*}

    \begin{table*}
    \centering
        \begin{tabular}{@{}l@{}c@{\,}|@{\ }c@{\ \ }c@{\ }c@{}}
        & & \multicolumn{3}{c}{ReVerb45K} \\
         Model & Pre-trained? & MR & MRR & H@10 \\ \hline
         \multirow{2}{*}{\footnotesize\textsc{NoEncoder\_TuckER}} & no &  $5110 (157.7)$ & $.096 (.002)$ & $.122 (.003)$ \\
          & yes & $\mathit{810 (6.04)}$ & $\mathit{.299 (.001)}$ & $\mathit{.451 (.002)}$ \\ \hline
         \multirow{2}{*}{\footnotesize\textsc{NoEncoder\_ConvE}} & no & $2837 (71.67)$ & $.223 (.001)$ & $.324 (.002)$ \\
          & yes & $\mathit{660 (13.54)}$ & $\mathit{.344 (.001)}$ & $\mathit{.502 (.001)}$ \\ \hline
         \multirow{2}{*}{\footnotesize\textsc{NoEncoder\_$5\star$E}} & no & $3176 (41.59)$ & $.151 (.001)$ & $.214 (.000)$ \\
          & yes & $\mathit{2723(20.15)}$ & $\mathit{.176(.000)}$ & $\mathit{.248 (.002)}$ \\ \hline
         \multirow{2}{*}{\footnotesize\textsc{GRU\_TuckER}} & no & $2047 (15.49)$ & $.267 (.003)$ & $.364 (.004)$ \\
          & yes & $\mathit{808 (75.13)}$ & $\mathit{.322 (.001)}$ & $\mathit{.458 (.008)}$ \\ \hline
         \multirow{2}{*}{\footnotesize\textsc{GRU\_ConvE}} & no & $904 (37.97)$ & $.335 (.004)$ & $.476 (.007)$ \\
          & yes & $\mathit{582 (19.25)}$ & $\mathit{.353 (.001)}$ & $\mathit{.507 (.004)}$ \\ \hline
         \multirow{2}{*}{\footnotesize\textsc{GRU\_$5\star$E}} & no &  $819 (30.51)$ & $.353 (.002)$ & $.511 (.002)$ \\ 
          & yes & $\boldsymbol{\mathit{572 (16.9)}}$ & $\boldsymbol{\mathit{.376 (.001)}}$ & $\boldsymbol{\mathit{.534 (.001)}}$ \\ \hline \hline
          $\blacklozenge$ \footnotesize\textsc{Okgit(ConvE)} & yes$^\blacktriangle$ & $773.9$ & $.332$ & $.464$ \\ 
          $\dagger$ \footnotesize\textsc{CaRe(ConvE, LAN)} & no & $1308$ & $.324$ & $.456$ \\ 
          $\dagger$ \footnotesize\textsc{TransE} & no & $2956$ & $.193$ & $.361$ \\ 
          $\dagger$ \footnotesize\textsc{TransH} & no & $2998$ & $.194$ & $.362$ \\ \hline \hline
            \multirow{2}{*}{\footnotesize\textsc{NoEncoder\_TuckER}\textsubscript{$d=200$}}\hspace{-2ex} & no & $5911 (53.85)$ & $.087 (.001)$ & $.107 (.002)$ \\
          & yes & $824 (12.74)$ & $.275 (.000)$ & $.423 (.001)$  \\ \hline
            \footnotesize\textsc{BigGraph\_TuckER}\textsubscript{$d=200$}\hspace{-2ex} & yes$^\blacktriangle$& $2285$ & $.234$ & $.337$ \\ \hline 
    \end{tabular}
        \caption{Comparison of different models with and without pre-training on the OKBC benchmark \textsc{ReVerb45K.}
        The scores of each model are reported with and without pre-training, with the better of the two written in \textit{italics}. Each score is averaged over 5 runs, with their standard deviation given in parentheses. Separated from the rest with two lines are the previous best results on the datasets, and \textsc{TuckER} results with $d=200$ for a fair \textsc{BigGraph} comparison. The best overall value in each column is written in \textbf{bold}.
    Results denoted with $\dagger$ and $\blacklozenge$ were taken from~\citep{gupta2019care} and~\citep{chandrahas2021okgit}, respectively.\\[1ex]
    $^\blacktriangle$ Unlike all other models with a \textit{yes} entry, \textsc{BigGraph\_TuckER} and \textsc{Okgit} were not pre-trained on \textsc{OlpBench}, but on pre-trained WikiData embeddings and with the masked language modeling objective, respectively.}
    \label{table:ReVerb45Results}
    \end{table*}
    The results of experiments on \textsc{ReVerb20k} and \textsc{ReVerb45k} are given in Tables~\ref{table:ReVerb20KResults} and~\ref{table:ReVerb45Results}.
    The models strictly improve their performance when pre-trained on \textsc{OlpBench}.
    This improvement is particularly noticeable for \textsc{NoEncoder} models, which tend to overfit and achieve poor results without pre-training.
    However, when initialized with a pre-trained model, their ability to generalize improves.
    \textsc{5$^\star$E} seems to be an exception to this, likely because there are no shared parameters between relations and entities, resulting in a weaker regularization.
    \textsc{Gru}-based models do not seem to suffer as severely from overfitting, but their performance still visibly improves when they are pre-trained on \textsc{OlpBench}.

    Finally, the best introduced model outperforms the state-of-the-art approach by \citet{chandrahas2021okgit} on \textsc{ReVerb20k} and \textsc{ReVerb45k}.
    Even when compared to pre-trained \textsc{Gru\_ConvE}, which is based on the same KBC model, \textsc{Okgit(ConvE)} and \textsc{CaRe(ConvE, Lan)} lag behind.
    This is particularly surprising, because the \textsc{Bert} and \textsc{RoBERTa} language models, used by \textsc{Okgit(ConvE)}, had received several orders of magnitude more pre-training on unstructured text, making the results of the introduced approach even more significant.

    Similarly, the initialization of models with \textsc{BigGraph} seems to help the performance, however, such an approach is in turn outperformed by a \textsc{NoEncoder} model, initialized with pre-trained encoders instead.
    This indicates that the suggested pre-training is much more efficient, despite the smaller computational cost.

\paragraph*{KBC results}

    \begin{table*}
    \centering
    \begin{tabular}{@{}l@{}c@{\ \ }|@{\ \ }c@{\ \ \ }c@{\ \ \ }c@{}}
        & & \multicolumn{3}{c}{\textsc{Fb15k237}} \\
     Model & pre-trained? & MR & MRR & H@10 \\ \hline
     \multirow{2}{*}{\footnotesize\textsc{TuckER}} & no & $203 (3.82)$ &$.351 (.001)$ & $.531 (.001)$ \\
      & yes & $\mathit{169 (1.52)}$ &$\mathit{.356 (.001)}$&$\mathit{.537 (.001)}$  \\ \hline
     \multirow{2}{*}{\footnotesize\textsc{ConvE}} & no & $164 (1.97)$ &$.332 (.001)$ & $\mathit{.518 (.002)}$ \\
      & yes & $\mathit{157 (0.77)}$ &$\mathit{.355 (.001)}$&$.509 (.001)$ \\ \hline 
     \multirow{2}{*}{\footnotesize\textsc{$5\star$E}} & no & $161 (0.25)$ &$.350 (.001)$&$.538 (.001)$ \\
      & yes & $\boldsymbol{\mathit{153 (1.15)}}$ &$\mathit{.355 (.001)}$&$\mathit{.541 (.0001)}$ \\ \hline \hline
        \textsuperscript{1} \footnotesize\textsc{ConvE} & no & $244$ & $.325$ & $.501$  \\
       \textsuperscript{2} \footnotesize\textsc{ConvE} & no & -- & $.339$ & $.536$  \\
       \textsuperscript{3} \footnotesize\textsc{TuckER} & no & -- & $.358$ & $.544$ \\ 
       \textsuperscript{4} \footnotesize\textsc{$5\star$E} & no & -- & $\boldsymbol{.37}$ & $\boldsymbol{.560}$ \\ \hline
       \textsuperscript{5} \footnotesize\textsc{KG-Bert} & yes & $\boldsymbol{153}$ & -- & $.420$  \\ 
       \textsuperscript{6} \footnotesize\textsc{BoxE} & no & $163$ & $.337$ & $.538$  \\ 
       \textsuperscript{7} \footnotesize\textsc{ComplEx} & no & -- & $\boldsymbol{.370}$ & $\boldsymbol{.560}$  \\
       \textsuperscript{8} \footnotesize\textsc{ComplEx-Dura} & no & -- & $\boldsymbol{.371}$ & $\boldsymbol{.560}$  \\
       \textsuperscript{8} \footnotesize\textsc{Rescal-Dura} & no & -- & $.368$ & $.550$ \\
       \textsuperscript{9} \footnotesize\textsc{MLMLM} & yes & 411 & $.259$ & $.403$ \\ \hline
    \end{tabular}
        \caption{Comparison of different models with and without pre-training on the KBC benchmarks \textsc{Fb15k237}. 
        The scores of each model are reported with and without pre-training, with the better of the two written in \textit{italics}. 
        Separated from the rest with two lines, previous scores obtained with the \textsc{ConvE}, \textsc{5$^\star$E}, and \textsc{TuckER} models are listed, followed by other well-performing models in the literature. 
        The best overall result in each column is highlighted in \textbf{bold}.\\[1ex]
        \textsuperscript{1}\citep{dettmers2018conve}; \textsuperscript{2}\citep{ruffineli2020olddog};\textsuperscript{3}\citep{balazevic2019tucker};\textsuperscript{4}\citep{nayyeri20215stare};\textsuperscript{5}\citep{yao2019kgbert};
        \textsuperscript{6}\citep{abboud2020boxe};\textsuperscript{7}\citep{lacroix2018complex};\textsuperscript{8}\citep{zhang2020dura};\textsuperscript{9}\citep{mlmlm}.
        }
    \label{table:fb15kResults}
    \end{table*}

       \begin{table*}
    \centering
    \begin{tabular}{@{}l@{}c@{\ \ }|@{\ \ }c@{\ \ \ }c@{\ \ \ }c@{}}
        & & \multicolumn{3}{c}{\textsc{Wn18rr}} \\
     Model & pre-trained? & MR & MRR & H@10 \\ \hline
     \multirow{2}{*}{\footnotesize\textsc{TuckER}} & no & $3801 (62.68)$ & $.466 (.001)$ & $.528 (.002)$ \\
      & yes &  $\mathit{3526 (90.86)}$ & $\mathit{.468 (.004)}$ & $\mathit{.529 (.003)}$ \\ \hline
     \multirow{2}{*}{\footnotesize\textsc{ConvE}} & no & $6187 (91.79)$ & $.425 (.002)$ & $.471 (.003)$ \\
      & yes & $\mathit{5811 (63.42)}$ & $\mathit{.427 (.002)}$ & $\mathit{.474 (.003)}$ \\ \hline 
     \multirow{2}{*}{\footnotesize\textsc{$5\star$E}} & no & $2686 (209.96)$ & $\mathit{.489 (.002)}$ & $.574 (.002)$ \\
      & yes & $\mathit{2674 (87.23)}$ & $\mathit{.491 (.001)}$ & $\mathit{.579 (.003)}$ \\ \hline \hline
        \textsuperscript{1} \footnotesize\textsc{ConvE} & no & $4187$ & $.430$ & $.520$ \\
       \textsuperscript{2} \footnotesize\textsc{ConvE} & no & -- & $.442$ & $.504$ \\
       \textsuperscript{4} \footnotesize\textsc{$5\star$E} & no & -- & $.500$ & $.590$ \\ \hline
       \textsuperscript{5} \footnotesize\textsc{KG-Bert} & yes & $\boldsymbol{97}$ & -- & $.524$ \\ 
       \textsuperscript{6} \footnotesize\textsc{BoxE} & no & $3207$ & $.451$ & $.541$ \\ 
       \textsuperscript{7} \footnotesize\textsc{ComplEx} & no & -- & $.480$ & $.570$ \\
       \textsuperscript{8} \footnotesize\textsc{ComplEx-Dura} & no  & -- & $.491$ & $.571$ \\
       \textsuperscript{8} \footnotesize\textsc{Rescal-Dura} & no & -- & $.498$ & $.577$ \\
       \textsuperscript{9} \footnotesize\textsc{MLMLM} & yes & 1603 & $\boldsymbol{.502}$ & $\boldsymbol{.611}$ \\ \hline
    \end{tabular}
        \caption{Comparison of different models with and without pre-training on the KBC benchmarks \textsc{Wn18rr}. 
        The scores of each model are reported with and without pre-training, with the better of the two written in \textit{italics}. 
        Separated from the rest with two lines, previous scores obtained with the \textsc{ConvE}, \textsc{5$^\star$E}, and \textsc{TuckER} models are listed, followed by other well-performing models in the literature. 
        The best overall result in each column is highlighted in \textbf{bold}.\\[1ex]
        \textsuperscript{1}\citep{dettmers2018conve}; \textsuperscript{2}\citep{ruffineli2020olddog};\textsuperscript{3}\citep{balazevic2019tucker};\textsuperscript{4}\citep{nayyeri20215stare};\textsuperscript{5}\citep{yao2019kgbert};
        \textsuperscript{6}\citep{abboud2020boxe};\textsuperscript{7}\citep{lacroix2018complex};\textsuperscript{8}\citep{zhang2020dura};\textsuperscript{9}\citep{mlmlm}.
        }
    \label{table:wn18Results}
    \end{table*}

    To evaluate the impact of pre-training on larger canonicalized knowledge bases, we compare the performance of models on \textsc{Fb15k237} and \textsc{Wn18rr}.
    For brevity, we treat the choice of an encoder as a hyperparameter and report the better of the two models in Tables~\ref{table:fb15kResults} and \ref{table:wn18Results}. 

    Pre-trained models outperform their randomly initialized counterparts as well, however, the differences are usually smaller.
    There are several reasons that can explain the small difference in performance, primarily the difference in dataset size.
    The best-performing models on \textsc{Fb15k237} and \textsc{Wn18rr} only made between $3$ and $12$ times more steps during pre-training than during fine-tuning.
    For comparison, this ratio was between $250$ to $1000$ for \textsc{ReVerb20k}.
    The smaller improvements on \textsc{Fb15k237} and \textsc{Wn18rr} can also be explained by the domain shift, as already described in Section~\ref{section:kbc-datasets}.

    Table~\ref{table:fb15kResults} additionally includes multiple recently published implementations of \textsc{ConvE, 5$^\star$E}, and \textsc{TuckER}, as well as other well-performing models in KBC.
    The comparison with all these models should be taken with a grain of salt, as other reported models were often trained with a much larger hyperparameter space, as well as additional techniques for regularization (e.g., \textsc{Dura}~\citep{zhang2020dura} or label smoothing~\citep{balazevic2019tucker}) and sampling (e.g., self-adversarial sampling~\citep{abboud2020boxe}).
    Due to a large number of models and baselines, we could not expand the hyperparameter search without compromising the fair evaluation of all compared models.

    The pre-trained models usually obtain their best result with a larger dimension compared to their randomly-initialized counterparts. 
    Pre-training thus serves as a type of regularization, allowing us to fine-tune larger models.
    For this reason, pre-training on even larger datasets and increasing the number of parameters is likely to result in even more significant improvements across many KBC datasets in future work.

    \begin{figure}[t]
        \centering
        \includegraphics[width=0.60\textwidth]{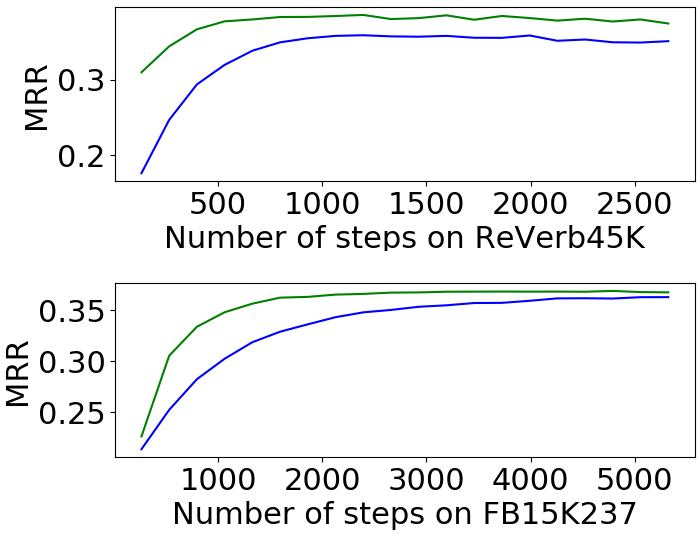}
        \caption{Comparing convergence of the best randomly initialized (blue) and pre-trained (green) models on \textsc{ReVerb45k} and \textsc{Fb15k237}. 
        Pre-trained models converge in fewer training steps despite a smaller learning rate.}
        \label{figure:convergence}
    \end{figure}
    Not only does pre-training allow training of larger models, the pre-trained models also require fewer training steps to obtain a similar performance, as seen in Figure~\ref{figure:convergence}.
    Even though the pre-trained models were fine-tuned with a smaller learning rate, they converge in fewer steps, which can be attributed to the pre-training.
    The experiments summarized in the figure were performed with the best hyperparameter setup for both pre-trained and randomly initialized models. 

    Finally, we highlight that language modeling as pre-training hardly justifies the enormous computational cost.
    On the majority of metrics, \textsc{KG-Bert} performs worse than models with fewer parameters and less pre-training, with a notable exception of remarkable MR on the \textsc{Wn18rr} dataset.

\subsection{Zero-Shot Experiments}
\label{section:kbc-zeroshot}

    \begin{table*}[ht] 
    \centering
        \begin{tabular}{@{}l@{}c@{\ }|@{\ }c@{\ }c@{\ }c@{\ }|@{\ }c@{\ }c@{\ }c@{}}
        & & \multicolumn{3}{c}{\textsc{ReVerb20k}} & \multicolumn{3}{c}{\textsc{ReVerb45k}} \\
     Model & pre-trained? & MR & MRR & Hits@10 & MR & MRR & Hits@10 \\ \hline
     \multirow{2}{*}{\footnotesize\textsc{NoEncoder\_TuckER}} & no & $4862$ &$.001$&$.002$ & $8409$ & $.001$ & $.001$ \\
      & yes & $\mathit{1486}$ &$\mathit{.012}$&$\mathit{.026}$ & $\mathit{1737}$ & $\mathit{.019}$ & $\mathit{.041}$ \\ \hline
     \multirow{2}{*}{\footnotesize\textsc{NoEncoder\_ConvE}} & no & $4600$ &$.002$&$\mathit{.004}$ & $8464$ & $.001$ & $\mathit{.001}$ \\
      & yes & $\mathit{1480}$ &$\mathit{.003}$&$.000$ & $\mathit{1962}$ & $\mathit{.003}$ & $\mathit{.001}$ \\ \hline
     \multirow{2}{*}{\footnotesize\textsc{NoEncoder\_$5\star$E}} & no & $4795$ &$.001$&$.000$ & $8445$ & $.001$ & $.002$ \\
      & yes & $\mathit{1422}$ &$\mathit{.005}$&$.000$ & $\mathit{1701}$ & $\mathit{.014}$ & $\mathit{.011}$ \\ \hline
    \end{tabular}
        \caption{Comparison of the zero-shot performance of different models on the OKBC benchmarks \textsc{ReVerb20k} and \textsc{ReVerb45k}. The scores of each model are reported with and without pre-training on \textsc{OlpBench}, with the better of the two written in \textit{italics}. Note that the model that was not pre-trained is equivalent to a random baseline.}
    \label{table:0shotResults}
    \end{table*}

    To gain a better understanding of what kind of knowledge is transferred between the datasets, we investigate the zero-shot performance of pre-trained models on \textsc{ReVerb20k} and \textsc{ReVerb45k}, where the impact of pre-training was the strongest.
    That means that we follow the same setup as in Section~\ref{Section:Doge-experiments} with all models.
    If pre-training improvement was mainly due to the direct memorization of facts, the zero-shot performance should already be high without fine-tuning.

    The results of the zero-shot evaluation are given in Table~\ref{table:0shotResults}.
    Given is the performance of all models with and without pre-training, the latter being equivalent to a random baseline.
    Note that since no fine-tuning takes place, the choice of the encoder for evaluation does not matter.
    We thus chose to only evaluate one of them.

    Observing the results, we see that pre-training the models results in a lower MR, but not necessarily in a much higher MRR.
    Even when MRR does improve, the difference is much smaller than when comparing fine-tuned models in Tables\ref{table:ReVerb20KResults} and ~\ref{table:ReVerb45Results}.
    This implies that the improvement induced by pre-training likely does not happen only due to the memorization of facts from \textsc{OlpBench}.
    On the other hand, the MR of pre-trained models is comparable or even better than the MR of randomly initialized \textsc{NoEncoder} models, fine-tuned on the \textsc{ReVerb} datasets, as reported in Tables~\ref{table:ReVerb20KResults} and ~\ref{table:ReVerb45Results}.
    Hence, pre-trained models carry a lot of ``approximate knowledge''.
    This means that correct answers tend to not be ranked low, even when the model does not give the correct answer.
    This is consistent with earlier remarks on pre-training serving as a type of regularization. Table~\ref{table.examples} shows several corresponding examples of the ConVE model with NoEncoder on the ReVerb20K dataset. It shows that the correct answers are ranked better in the pre-trained model. These are all instances of ``general knowledge" that are hard to deduce from the training set alone---e.g., ``crude oil" appears only once in the training set of \textsc{ReVerb20K}. On the other hand, we can find the fact $\langle\text{crude oil, is used as, fossil fuel}\rangle$ in \textsc{OlpBench}, from which the first fact can be deduced.

    Knowing that the \textsc{ReVerb20k} and \textsc{ReVerb45k} test sets consist of facts which contain at least one previously unseen entity or relation, this experiment can be viewed as a test of out-of-distribution generalization.
    Comparing the zero-shot MRR results with the \textsc{OlpBench} results suggests that while OKBC models are capable of out-of-domain generalization to unseen entities and relations, there is still room for improvement.


\begin{table*}[ht]
	\begin{center}
		\renewcommand{\arraystretch}{1.0}
		\footnotesize{
			\centering{\setlength\tabcolsep{3pt}
				\begin{tabular}{c|c|c}
					\toprule
                    Data Instance & Rank (\textsc{no}) & Rank (\textsc{yes}) \\ \midrule
                    
                    \makecell{\textsc{Head}: fuel, \textsc{Relation}: be make from, \textsc{Tail}: \underline{crude oil} \\ 
                    \textsc{Wrong Answers}: constipation, vegetable, 1831, heaven, \\ hour, high school,	original, dmca, version} & 5 & 1 \\ \midrule

                    \makecell{\textsc{Head}: \underline{ethanol fuel}, \textsc{Relation}: be made of, \textsc{Tail}: ethanol \\ 
                    \textsc{Wrong Answers}: population genetic,	hannibal, tesla roadste, \\dvd, identity theft, corey feldman, Clinton, gate, monroe} & 6 & 2 \\ \midrule    

                    \makecell{\textsc{Head}: nyx, \textsc{Relation}: be the goddess of, \textsc{Tail}: \underline{darkness} \\ 
                    \textsc{Wrong Answers}: united states, presidency, sale, Russia, \\bad idea, king, progress, obama, event} & 9 & 1 \\ \bottomrule
                  
		\end{tabular}}}
	\vspace{-2ex}
	\caption{Examples of predictions of ConVE model with NoEncoder on ReVerb20K dataset. \textsc{yes} and \textsc{no} denote whether we used pre-trained model or not. The word underlined is the target answer. Wrong answers are randomly sampled from entities listed in the test set.} \label{table.examples}%
	\end{center}
    \vspace{-3.5ex}
\end{table*}

\section{Results of Diagnostic Experiments}
\label{Section:Doge-results}
    This chapter contains the results of all evaluated models on all parts of the \textsc{Doge} dataset.
    Results are reported in MR, MRR, and Hits@1 metrics, except for parts of the dataset that measure consistency, for which the mean and standard deviation are reported instead.
    Hits@N metrics for other values of N were not reported due to the small number of possible answers ($10$ or~$50$).

\subsection{General Knowledge}

    All three pre-trained KBC architectures are compared on the general knowledge part of the \textsc{Doge} dataset, with results reported in Table~\ref{table:GeneralKnowledge}.
    It is evident that the models are fairly consistent with each other relative to how well they do in different categories.
    This is unsurprising since they were all trained on \textsc{OlpBench}.
    Nevertheless, it also means that it can be difficult to conclude which of the models handles what type of data better.
    \textsc{ConvE} seems to often outperform other models in most categories, despite often ranking the lowest of the three models on commonly used KBC benchmarks.
    
    Inconsistent results between \textsc{Doge} and most other datasets can be explained by the difference in their construction.
    Unlike other KBC datasets, \textsc{Doge} was not obtained by randomly splitting an existing knowledge base.
    Models trained on \textsc{OlpBench} and evaluated on \textsc{Doge} face a shift of data distribution, requiring better generalization capabilities.
    This is the first result of this kind, indicating that existing KBC benchmarks only cover a specific scenario, which may incorrectly reward models with lower generalization capabilities.

    \begin{table}[th]
        \centering
        \begin{tabular}{@{}l@{\ }|@{\ }c@{\ }c@{\ }c@{\ }|@{\ }c@{\ }c@{\ }c@{\ }|@{\ }c@{\ }c@{\ }c@{}}
             & \multicolumn{3}{c}{\textsc{TuckER}} & \multicolumn{3}{c}{\textsc{ConvE}} & \multicolumn{3}{c}{\textsc{5$^\star$E}} \\
            \footnotesize category & MR & MRR & H@1 & MR & MRR & H@1 & MR & MRR & H@1 \\ \hline
            \footnotesize Entertain. \& Pop Culture & $3.55$ & $.510$ & $.31$ & $3.49$ & $.520$ & $.33$ & $3.40$ & $.559$ & $.39$ \\ 
            \footnotesize Geography \& Travel & $3.03$ & $.601$ & $.43$ & $2.87$ & $.620$ & $.46$ & $2.40$ & $.702$ & $.57$\\
            \footnotesize Health \& Medicine & $3.95$ & $.461$ & $.27$ & $3.43$ & $.493$ & $.27$ & $3.18$ & $.561$ & $.39$ \\
            \footnotesize Lifestyle \& Social Issues & $3.35$ & $.535$ & $.36$ & $3.21$ & $.564$ & $.39$ & $3.25$ & $.567$ & $.39$ \\
            \footnotesize Literature & $3.69$ & $.509$ & $.33$ & $3.30$ & $.550$ & $.37$ & $3.08$ & $.602$ & $.45$ \\
            \footnotesize Philosophy \& Religion & $3.22$ & $.545$ & $.37$ & $3.25$ & $.544$ & $.34$ & $3.01$ & $.599$ & $.43$ \\
            \footnotesize Politics, Law \& Government & $3.45$ & $.532$ & $.35$ & $2.92$ & $.563$ & $.36$ & $2.79$ & $.639$ & $.50$\\
            \footnotesize Science & $3.60$ & $.505$ & $.32$ & $3.94$ & $.450$ & $.23$ & $3.28$ & $.569$ & $.41$\\
            \footnotesize Sports \& Recreation & $3.71$ & $.470$ & $.25$ & $3.63$ & $.472$ & $.27$ & $3.02$ & $.587$ & $.41$\\
            \footnotesize Technology & $3.45$ & $.532$ & $.35$ & $3.30$ & $.534$ & $.34$ & $3.13$ & $.593$ & $.43$\\
            \footnotesize Visual Arts & $3.77$ & $.480$ & $.28$ & $3.66$ & $.492$ & $.32$ & $3.50$ & $.557$ & $.40$\\
            \footnotesize World History & $3.28$ & $.520$ & $.30$ & $3.19$ & $.580$ & $.42$ & $2.55$ & $.642$ & $.46$
        \end{tabular}
        \caption{Comparison of the three pre-trained architectures on the general knowledge subset of the \textsc{Doge} dataset.
        Each column contains scores of a model on different categories, as defined by Encyclopedia Britannica.}
        \label{table:GeneralKnowledge}
    \end{table}

\subsection{Model Consistency}

    All three KBC pre-trained architectures are compared on their ability to make consistent prediction in the presence of synonyms and inverse relations, with results reported in Table~\ref{table:consistency}.
    None of the models are particularly consistent on either of the tests, with \textsc{ConvE} being worse than \textsc{5$^\star$E}, especially in ``Relation Synonyms''. The discrepancy in consistency may  be due to the comparably better performance of \textsc{5$^\star$E} on \textsc{Doge}, particularly in ``Science'' and ``World History'' topics, where the annotated data points for ``Relation Synonyms'' account for 19\% of total data. Overall, it is safe to conclude that none of the tested models can be called \textit{consistent}, which is unfortunately common for neural network models.


    \begin{table}[th]
        \centering
        \begin{tabular}{@{}l@{\ }|cc|cc|cc@{}}
             & \multicolumn{2}{c}{\textsc{TuckER}} & \multicolumn{2}{c}{\textsc{ConvE}} & \multicolumn{2}{c}{\textsc{5$^\star$E}} \\
             & mean & stdev & mean & stdev& mean & stdev \\ \hline
            \footnotesize Entity Synonyms & $0.692$ & $3.910$ & $0.931$ & $3.397$ & $0.836$& $3.559$ \\ 
            \footnotesize Relation Synonyms & $0.177$ & $3.027$ & $0.403$ & $2.755$ & $0.109$ & $1.958$ \\
            \footnotesize Inverse Relations & $0.239$ & $3.030$ & $0.183$ & $2.609$ & $0.090$ & $2.119$ \\
        \end{tabular}
        \caption{A comparison of three pre-trained architectures on subsets of the \textsc{Doge} dataset that probe the consistency of predictions.
        Mean values are usually close to zero, indicating that the deviation was not caused by poor annotation.}
        \label{table:consistency}
    \end{table}

\subsection{Deductive Reasoning}
    
    In this section, we report the results on the test of deductive reasoning.
    The scores obtained by all three evaluated models are given in Table~\ref{table:deductive}.
    All models achieve approximately chance performance on the \textit{No Added Facts} experiments, indicating that the answers of this subset cannot be deduced through undesired clues.
    Moreover, all models achieve a better-than-random guessing score on the \textit{Background Knowledge} experiments, indicating that they do contain the relevant background knowledge for many instances.
    If the results of the models on this set of instances were poor, the results from this whole section would be inconclusive.
    Finally, the results on \textit{With Added Facts} lie somewhere in between the two, as expected.
    This indicates that the models can indeed combine newly obtained facts with their background knowledge, however, a big gap to \textit{Background Knowledge} results indicates that there is a lot of room for improvement.

    The impact of the encoder varies by KBC model.
    For \textsc{ConvE} and \textsc{TuckER}, there is no conclusive answer to which encoder is the better choice, as it depends on the choice of metric.
    \textsc{NoEncoder} versions obtain better MR, while \textsc{Gru} versions obtain better MRR and Hits@1 scores.
    For \textsc{5$^\star$E}, \textsc{NoEncoder} is visibly the superior choice, making the \textsc{NoEncoder\_5$^\star$E} the best model for deductive reasoning out of all the compared setups.
    The reason behind this likely lies in its shallow architecture.
    Since there are no shared parameters or stacked layers, training primarily affects the embeddings of entities in the training data while preserving the structure of everything else.


    \begin{table}[th]
        \centering
        \begin{tabular}{@{}l@{\ }|@{\ }c@{\ }c@{\ }c@{\ }|@{\ }c@{\ }c@{\ }c@{\ }|@{\ }c@{\ }c@{\ }c@{}}
             & \multicolumn{3}{c}{\textsc{TuckER}} & \multicolumn{3}{c}{\textsc{ConvE}} & \multicolumn{3}{c}{\textsc{5$^\star$E}} \\
             & MR & MRR & H@1 & MR & MRR & H@1 & MR & MRR & H@1 \\ \hline
            \footnotesize Background Knowledge & $3.26$ & $.556$ & $.373$ & $3.01$ & $.598$ & $.435$ & $3.24$ & $.574$ & $.408$\\
            \footnotesize No Added Facts & $5.09$ & $.343$ & $.15$ & $5.07$ & $.334$ & $.13$ & $4.65$ & $.350$ & $.14$\\
            \footnotesize With Added Facts\textsubscript{\textsc{Gru}} & $4.16$ & $.457$ & $.28$ & $4.33$ & $.403$ & $.22$ & $4.11$ & $.464$ & $.30$ \\
            \footnotesize With Added Facts\textsubscript{\textsc{NoEnc.}} & $4.39$ & $.415$ & $.23$ & $4.69$ & $.344$ & $.13$ & $4.25$ & $.416$ & $.22$ \\
        \end{tabular}
        \caption{Comparison of the three pre-trained architectures on the deductive reasoning subset of the \textsc{Doge} dataset.
        \textit{Background Knowledge} and \textit{No Added Facts} results serve as soft upper and lower bounds to the \textit{With Added Facts} results, which can be correctly deduced only by combining newly added facts with the relevant background knowledge.
        Fine-tuning is done both with \textsc{Gru} and \textsc{NoEncoder} setup to compare the impact of the two.}
        \label{table:deductive}
    \end{table}

\subsection{Gender Stereotypes}

    Finally, all deductive reasoning tests are repeated on the subset of \textsc{Doge} that detects gender stereotypes.
    The results are given in Table~\ref{table:Stereotypes} and are reported separately for each gender and separately for stereotypical and anti-stereotypical occupations.
    Finally, we test whether the model predictions are affected by the stereotypes indicating the presence of historical bias.
    To see whether the difference in performance is statistically significant, we use the Wilcoxon signed-rank test.
    More specifically, we test whether swapping the name of a gender that is not stereotypically associated with an occupation for a name of the opposite gender impacts the rank of the correct answer.

    \begin{table}[ht!]
        \centering
            \begin{footnotesize}

        \begin{tabular}{p{3.7cm}|p{0.65cm}p{0.65cm}p{0.55cm}|p{0.65cm}p{0.65cm}p{0.55cm}|p{0.65cm}p{0.65cm}p{0.55cm}}
             & \multicolumn{3}{c}{\textsc{TuckER}} & \multicolumn{3}{c}{\textsc{ConvE}} & \multicolumn{3}{c}{\textsc{5$^\star$E}} \\
            \footnotesize No Added Facts & MR & MRR & H@1 & MR & MRR & H@1 & MR & MRR & H@1 \\ \hline
            \footnotesize St Masculine & $22.51$ & $.096$ & $.02$ & $24.27$ & $.083$ & $.02$ & $22.9$ & $.096$ & $.01$ \\ 
            \footnotesize St Feminine & $21.07$ & $.093$ & $.00$ & $18.45$ & $.138$ & $.02$ & $20.95$ & $.119$ & $.04$\\
            \footnotesize Anti-St Masculine & $29.96$ & $.079$ & $.03$ & $32.18$ & $.037$ & $.00$ & $30.32$ & $.066$ & $.01$ \\
            \footnotesize Anti-St Feminine & $29.21$ & $.062$ & $.00$ & $28.41$ & $.0642$ & $.00$ & $28.11$ & $.072$ & $.02$ \\ 
            \footnotesize p-value & \multicolumn{3}{c|}{$6.04\cdot10^{-15}$} & \multicolumn{3}{c|}{$2.19\cdot10^{-15}$}& \multicolumn{3}{c}{$1.48\cdot10^{-11}$} \\ \hline
            \footnotesize Background Knowledge & $9.575$ & $.417$ & $.275$ & $7.910$ & $.4307$ & $.265$ & $7.155$ & $.503$ & $.345$ \\ \hline
        \end{tabular}
        \vspace{5mm}\\
        \begin{tabular}{p{3.7cm}|p{0.65cm}p{0.65cm}p{0.55cm}|p{0.65cm}p{0.65cm}p{0.55cm}|p{0.65cm}p{0.65cm}p{0.55cm}}
             & \multicolumn{3}{c}{\textsc{TuckER}} & \multicolumn{3}{c}{\textsc{ConvE}} & \multicolumn{3}{c}{\textsc{5$^\star$E}} \\
            \footnotesize With Added Facts\textsubscript{\textsc{Gru}} & MR & MRR & H@1 & MR & MRR & H@1 & MR & MRR & H@1 \\ \hline
            \footnotesize St Masculine & $18.09$ & $.154$ & $.08$ & $18.69$ & $.176$ & $.09$ & $14.58$ & $.184$ & $.04$ \\ 
            \footnotesize St Feminine & $20.61$ & $.138$ & $.03$ & $18.22$ & $.224$ & $.14$ & $15.33$ & $.175$ & $.04$\\
            \footnotesize Anti-St Masculine & $24.37$ & $.084$ & $.01$ & $25.22$ & $.079$ & $.00$ & $19.16$ & $.198$ & $.10$ \\
            \footnotesize Anti-St Feminine & $26.83$ & $.109$ & $.03$ & $25.63$ & $.098$ & $.03$ & $21.36$ & $.119$ & $.03$ \\
            \footnotesize p-value & \multicolumn{3}{c|}{$7.20\cdot10^{-9}$} & \multicolumn{3}{c|}{$2.62\cdot10^{-12}$}& \multicolumn{3}{c}{$8.87\cdot10^{-6}$} \\ \hline
        \end{tabular}
        \vspace{5mm}\\
        \begin{tabular}{p{3.7cm}|p{0.65cm}p{0.65cm}p{0.55cm}|p{0.65cm}p{0.65cm}p{0.55cm}|p{0.65cm}p{0.65cm}p{0.55cm}}
            & \multicolumn{3}{c}{\textsc{TuckER}} & \multicolumn{3}{c}{\textsc{ConvE}} & \multicolumn{3}{c}{\textsc{5$^\star$E}} \\
            \footnotesize With Added Facts\textsubscript{\textsc{NoEnc.}} & MR & MRR & H@1 & MR & MRR & H@1 & MR & MRR & H@1 \\ \hline
            \footnotesize St Masculine & $16.68$ & $.168$ & $.07$ & $15.68$ & $.210$ & $.07$ & $21.72$ & $.114$ & $.02$ \\ 
            \footnotesize St Feminine & $20.23$ & $.125$ & $.00$ & $16.20$ & $.210$ & $.08$ & $22.05$ & $.111$ & $.03$\\
            \footnotesize Anti-St Masculine & $23.10$ & $.098$ & $.03$ & $21.36$ & $.099$ & $.01$ & $28.47$ & $.073$ & $.01$ \\
            \footnotesize Anti-St Feminine & $27.57$ & $.092$ & $.03$ & $22.58$ & $.1343$ & $.05$ & $26.55$ & $.073$ & $.01$ \\
            \footnotesize p-value & \multicolumn{3}{c|}{$7.03\cdot10^{-13}$} & \multicolumn{3}{c|}{$3.81\cdot10^{-10}$}& \multicolumn{3}{c}{$5.43\cdot10^{-9}$} \\ \hline
        \end{tabular}
\end{footnotesize}
        \caption{Comparison of three pre-trained architectures on the gender-bias detection subset of \textsc{Doge}.
        The dataset is split into Stereotypical (St) and Anti-Stereotypical (Anti-St) examples.
        The first of the three tables contains information on the experiments that do not require fine-tuning: stereotype detection and background knowledge.
        The second and the third table contain information on deductive reasoning about gender for \textsc{Gru} and \textsc{NoEncoder} encoders, respectively.}
        \label{table:Stereotypes}
    \end{table}

    The gender stereotypes are strongly present, resulting in a model being correct more often when the gender of the name matches the conceptual gender of the occupation.
    By comparing the results across all examples, both stereotypical and anti-stereotypical, we can see that the masculine names are on average ranked higher, likely due to representation bias in the training data.
    Moreover, all models obtain decent scores on background knowledge about occupations, making deductive reasoning experiments conclusive.

    The observed patterns persist when additional relevant data are  added to the models, regardless of the choice of the encoder.
    It is worth noting that the improvement from the additional knowledge about entities only moderately helps the models regardless of the gender or stereotypes.
    Connecting this with the results on deductive reasoning, we can conclude that models contain gender stereotypes and are unable to overcome them in the presence of additional data.
    The reason why they cannot overcome them lies in their general poor ability to perform deductive reasoning rather than a strong presence of stereotypes.
    Unfortunately, this means that all analyzed pre-trained KBC models can introduce gender-stereotyped facts into the target knowledge base, even in the presence of counter-evidence.

\subsection{The Impact of Word Embeddings}
    
    Finally, we report the results of re-training \textsc{ConvE} and \textsc{TuckER} on \textsc{OlpBench} without the \textsc{GloVe} word embeddings.
    The results on \textsc{OlpBench} are reported in Table~\ref{table:OlpBenchGlove}.
    Evidently, the absence of \textsc{GloVe} vectors significantly decreases the performance of both models.

    \begin{table}[ht!]
    \centering
    \begin{tabular}{l|ccc}
     & MR & MRR & H@10 \\ \hline
         \textsc{Gru\_TuckER\textsubscript{GloVe}} & $\mathit{57.2K}$ &$\mathit{.053}$&$\mathit{.097}$ \\ 
         \textsc{Gru\_TuckER\textsubscript{No\_GloVe}} & $65.1K$ &$.042$&$.077$ \\ \hline
         \textsc{Gru\_ConvE\textsubscript{GloVe}} & $\mathit{57.2K}$ & $\mathit{.045}$ & $\mathit{.086}$\\ 
         \textsc{Gru\_ConvE\textsubscript{No\_GloVe}} & $69.2K$ & $.030$ & $.057$\\ 
    \end{tabular}
        \caption{Comparison of \textsc{GloVe} impact on \textsc{OlpBench}. \textsc{TuckER} and \textsc{ConvE} are both trained with \textsc{GloVe} and with randomly-initialized word embeddings. 
        The better of the two in each column is written in \textit{italics}.}
    \label{table:OlpBenchGlove}
    \end{table}

    To see whether models without \textsc{GloVe} embeddings are less susceptible to gender stereotypes, we evaluate them on the gender stereotype subset of the \textsc{Doge} dataset.
    The results can be found in Table~\ref{table:StereotypesGlove}.
    Despite the slightly lower performance, the general trends in the results of these experiments are very similar to the results with the \textsc{GloVe} embeddings in Table~\ref{table:Stereotypes}.
    This indicates that \textsc{GloVe} embeddings definitely are not the only source of gender stereotypes, and while their removal results in an undesired performance drop, it does not decrease the biased behavior.

    \begin{table}[ht!]
        \centering
        \begin{tabular}{@{}l|ccc|ccc@{}}
             & \multicolumn{3}{c}{\textsc{TuckER}} & \multicolumn{3}{c}{\textsc{ConvE}} \\
            \footnotesize No Added Facts & MR & MRR & H@1 & MR & MRR & H@1 \\ \hline
            \footnotesize St Masculine & $23.34$ & $.071$ & $.00$ & $23.32$ & $.109$ & $.03$ \\ 
            \footnotesize St Feminine & $20.56$ & $.1396$ & $.05$ & $23.40$ & $.113$ & $.03$ \\
            \footnotesize Anti-St Masculine & $31.14$ & $.042$ & $.00$ & $27.80$ & $.054$ & $.00$\\
            \footnotesize Anti-St Feminine & $27.97$ & $.104$ & $.05$ & $27.70$ & $.070$ & $.00$ \\
            \footnotesize p-value & \multicolumn{3}{c|}{$3.03\cdot10^{-16}$} & \multicolumn{3}{c}{$1.67\cdot10^{-12}$} \\ \hline
            \footnotesize Background Knowledge & $10.19$ & $.342$ & $.200$ & $10.70$ & $.338$ & $.195$ \\ \hline
        \end{tabular}
        \vspace{5mm}\\
        \begin{tabular}{@{}l|ccc|ccc@{}}
             & \multicolumn{3}{c}{\textsc{TuckER}} & \multicolumn{3}{c}{\textsc{ConvE}} \\
            \footnotesize With Added Facts\textsubscript{\textsc{Gru}} & MR & MRR & H@1 & MR & MRR & H@1 \\ \hline
            \footnotesize St Masculine & $19.54$ & $.113$ & $.01$ & $19.12$ & $.170$ & $.07$ \\ 
            \footnotesize St Feminine & $21.21$ & $.132$ & $.04$ & $21.94$ & $.149$ & $.05$ \\
            \footnotesize Anti-St Masculine & $25.47$ & $.078$ & $.02$ & $23.39$ & $.099$ & $.02$\\
            \footnotesize Anti-St Feminine & $26.80$ & $.113$ & $.04$ & $26.23$ & $.088$ & $.02$\\
            \footnotesize p-value & \multicolumn{3}{c|}{$9.59\cdot10^{-10}$} & \multicolumn{3}{c}{$3.51\cdot10^{-07}$}\\ \hline
        \end{tabular}

        \caption{Comparison of pre-trained \textsc{ConvE} and \textsc{TuckER} architectures without \textsc{GloVe} embeddings on the gender-bias detection subset of \textsc{Doge}.
        The dataset is split into a dataset with Stereotypical (St) and one with Anti-Stereotypical (Anti-St) examples.
        The first of the two tables contains information on experiments that do not require fine-tuning: stereotype detection and background knowledge.
        The second of the two tables contains the information on deductive reasoning about gender for \textsc{Gru} encoders.
        Fine-tuning with the \textsc{NoEncoder} setup was omitted for brevity.}
        \label{table:StereotypesGlove}
    \end{table}
 
\section{Related Work}
\label{Section:Related-work}

This section discusses the related work in knowledge base completion, transfer learning in NLP, and existing approaches to using external data or higher-level structures in knowledge base completion.
The most attention is put on the discussion of the last part due to its closeness to this work.

\paragraph*{Review of knowledge base completion approaches}
Approaches to knowledge base completion can be roughly split into two types, translational models and semantic similarity models.
Translational models, such as TransE~\citep{lin2015ptranse}, BoxE~\citep{abboud2020boxe}, $5^\star$E~\citep{nayyeri20215stare}, or RotatE~\citep{sun2018rotate} usually model triplets as translations or similar geometric operations (e.g., rotations) in space.
They are commonly ``shallow'' in the sense that they do not contain any neural networks with multiple layers. Instead, their design is guided by mathematical intuition.

Semantic similarity models usually map the pair (head, relation) into a vector space and measure its semantic similarity (usually dot product) with possible tails. Examples of such models are TuckER~\citep{balazevic2019tucker}, ConvE~\citep{dettmers2018conve}, and ComplEx~\citep{trouillon2017complex}.
Unlike ConvE, which contains a convolutional neural network, TuckER and ComplEx are examples of bilinear models --- a special case of semantic similarity models. In such models, the score of a triplet can be obtained as a multiplication of a relation-specific matrix with vector embeddings of the head and the tail.

Not all models for knowledge base completion fall strictly into these two categories. KG-\textsc{Bert}~\citep{yao2019kgbert}, for example, takes the entire triplet as the input and does not compute any intermediate representations.

We decided not to discuss the performance of the mentioned models on benchmarks as part of the literature review.
Their accuracy strongly depends on the training regime, choice of the loss function, and other details, which are not consistent across their respective papers.
Surveys of existing KBC methods have repeatedly shown that older models can outperform more recent approaches when trained in the same environment~\citep{ruffineli2020olddog,ali2020pykeen}.

\paragraph*{Review of transfer learning in natural language processing}
Any model claiming the ability to process human language requires an immense amount of background knowledge of words and their meanings. 
All models operating on human language share vocabulary, prompting the development of common resources such as pre-trained models or word embeddings trained on large unsupervised corpora.
Pre-training of models is done by initializing the model weights, word embeddings, or other parts of a model on a separate task with an abundance of data~\citep{pennington2014glove, devlin2019BERT}.

While pre-training on supervised tasks is possible and sometimes even more sample-efficient~\cite{conneau2017infersent}, the accessibility of raw text makes unsupervised approaches the more scalable option, typically resulting in better performance overall.
A common existing approach to pre-training is to train a language model --- a computational model that assigns a probability score to the input sequence of words~\cite{bengio2003languagemodel} --- on a large corpus of raw text~\cite{howard2018ulmfit, brown2020gpt3}.
A conceptually similar but more common approach is masked language modeling~\cite{devlin2019BERT, liu2019roberta}, which is not language modelling in the strict sense, because it only assigns probabilities to individual missing (masked) tokens instead of the entire sequence.

As an example of a pre-training method that is not based on predicting missing pieces of text, we highlight the unsupervised multi-lingual paraphrasing approach by~\citet{lewis2020marge}.
They show that one can obtain a model for machine translation in an unsupervised manner by training to paraphrase articles about the same news.

Further discussion of recent advances in pre-training in NLP is beyond the scope of this paper.
For a more exhaustive survey, we direct the readers to~\citet{qiu2020pretrainingsurvey}

\paragraph*{Review of uses of external data in knowledge base completion}
While large-scale transfer learning in KBC is not yet a widely-used method, multiple attempts to incorporate higher-level structures or external knowledge exist.
The most common approach is to impose a hierarchy between entities or relations.
This information is already available inside the KB, and many approaches just adapt the scoring function to raise its importance.
\textit{CTransR} \citep{huang2016transr} collects the input data into clusters and learns a separate relation embedding for each cluster.
\textit{Semantically smooth embedding (SSE)} \citep{guo2015sse} and \textit{type-embodied knowledge representation (TKRL)} \citep{xie2016tkrl} assign types to entities and require the entities of the same type (determined through the \texttt{IsA} relation) to stay close together.

A different approach to incorporate additional information is to analyze paths.
Unlike regular approaches that only analyze single relations (paths of length $1$), models like \textit{PTransE} \citep{lin2015ptranse} additionally incorporate information on the paths of longer lengths.
The number of paths increases exponentially with the length, hence all such approaches resort to sampling and approximate methods.
Incorporating paths improved the quality of embeddings at the expense of computational complexity.
Thus, they do not appear in combination with more complex models.

The models described so far rely on the information provided by the knowledge base.
However, a lot of knowledge is not incorporated in the knowledge bases directly.
Various models try to increase the amount of available information by using human-written text.
The simplest approach of incorporating natural text is initialization of parameters from precomputed word embeddings based on the textual description of the entities.
This approach is used by the \textit{NTN} model \citep{socher2013ntn}.

More advanced models, such as \textit{DKRL} \citep{xie2016dkrl} and \textit{TEKE} \citep{wang2016teke}, train the entity embeddings jointly with the text model and try to align them.
\citet{wang2016teke} also define a textual context embedding and try to learn the mapping between the textual and knowledge base embedding.
\citet{xie2016dkrl}, on the other hand, make use of an encoder that generates an embedding given a description of the entity. They show that their approach generalizes even to previously unseen entities when a description of these entities is available.
Models that additionally incorporate natural text perform better when faced with previously unseen entities.
\citet{yao2019kgbert} make use of the large-scale pre-trained transformer model \textsc{Bert} to classify whether a fact is true.
The main drawback of such an approach is its speed -- \textsc{Bert} is large, and it can take weeks of GPU runtime to evaluate such a model on a benchmark test set.

Many of the ideas described in this section thus far have been later used to tackle open knowledge bases.
Earlier attempts at open knowledge base completion are tied to existing work on the canonicalization of knowledge bases.
To canonicalize open knowledge bases, automatic canonicalization tools cluster entities using manually defined features~\citep{galarraga2014canonicalizingOKB} or by finding additional information from external knowledge sources~\citep{vashishth2018cesi}.
\citet{gupta2019care} use clusters obtained with these tools to augment entity embeddings for KBC.
Note that~\citet{gupta2019care} use RNN-based encoders to encode relations, but not to encode entities.
\citet{broscheit2020olpbench}, on the other hand, introduce a model with RNN-based encoders for both entities and relations. 
Finally, \citet{chandrahas2021okgit} use both KB canonicalization tools and the large-scale pre-trained model \textsc{Bert}, combining their predictions to make a more informed decision.

\section{Conclusion}
\label{Section:Conclusion}
    In this work, we have introduced a novel approach to transfer learning or domain adaptation between various knowledge base completion datasets and introduced \textsc{Doge}, a novel diagnostic dataset for the analysis of pre-trained models for knowledge base completion. 
    The main strength of the introduced pre-trained method is the ability to benefit from pre-training on uncanonicalized knowledge bases, constructed from facts that were collected from unstructured text.
    Scaling the introduced method up would let us train large-scale pre-trained models, which have already shown to be incredibly successful in NLP.
    The proposed method was tested on $5$ different existing datasets, $1$ for pre-training and $4$ for fine-tuning, demonstrating that pre-training improves the performance of models.
    Pre-training turned out to be particularly beneficial on small-scale datasets, where the most significant gains were obtained, i.e., a $6\%$ absolute increase of MRR and a $65\%$ decrease of MR in comparison to the previously best method on \textsc{ReVerb20k}, despite not relying on large pre-trained models like \textsc{Bert}.
    
    We used the introduced \textsc{Doge} dataset to evaluate pre-trained models introduced in Section~\ref{Section:pre-training-experiments}.
    The results indicated that the existing pre-trained models lack robustness against synonyms and inverse relations and strongly rely on gender stereotypes.
    On a more positive note, they seem to cover all major areas of knowledge decently.
    The main highlighted problem with the obtained pre-trained models is the persistence of harmful stereotypes, which the model cannot remove even with the introduction of counter-evidence.
    It may thus happen that a pre-trained model includes an incorrect fact $\langle$Mary, is, childcare worker$\rangle$ into the target knowledge base even if the knowledge base contains the fact $\langle$Mary, repairs, cars$\rangle$.

    While this research rigorously investigated a specific direction of pre-training, i.e., training on an uncanonicalized knowledge base and transferring to smaller closed and open knowledge bases, there is a lot of potential on training on canonicalized knowledge bases as well as transferring to different related tasks, such as named entity recognition and knowledge base canonicalization.
    
    This work carries important implications for future developments of the field. Pre-training has led to important breakthroughs in other areas where neural networks are applied, and it is reasonable to expect that the same will happen in knowledge base completion.
    Scaling up the pre-training is expected to improve the performance of KBC models even further, allowing their use on small and scarce knowledge bases. This allows for their use on early-stage knowledge bases, simplifying their construction.
    
    Results on the \textsc{Doge} dataset demonstrate that existing KBC benchmarks do not test all properties exhaustively.
    This benchmark allows engineers to diagnose the behavior of their models more precisely, providing insights beyond the theoretical limitation and overall performance.
    
\section*{Acknowledgments}
This is a substantially extended and improved version of a preliminary paper at EMNLP 2021~\cite{kocijan2021KBCtransfer}.
The authors would like to thank Ralph Abboud for his helpful comments on the paper manuscript and Ana \v{S}tuhec for suggesting the dataset name. This work was supported by the
Alan Turing Institute under the EPSRC grant
EP/N510129/1, the AXA Research Fund, the
ESRC grant ``Unlocking the Potential of AI
for English Law'', and the EPSRC Studentship
OUCS/EPSRC-NPIF/VK/1123106. We also acknowledge the use of the EPSRC-funded Tier 2
facility JADE (EP/P020275/1) and GPU computing support by Scan Computers International Ltd. 

\bibliographystyle{elsarticle-num-names-alpha}

\bibliography{refs.bib}
\end{document}